\documentclass[sigconf]{acmart}

\AtBeginDocument{%
  \providecommand\BibTeX{{%
    \normalfont B\kern-0.5em{\scshape i\kern-0.25em b}\kern-0.8em\TeX}}}

\copyrightyear{2024}
\acmYear{2024}
\setcopyright{rightsretained}
\acmConference[KDD '24]{Proceedings of the 30th ACM SIGKDD Conference on Knowledge Discovery and Data Mining}{August 25--29, 2024}{Barcelona, Spain}
\acmBooktitle{Proceedings of the 30th ACM SIGKDD Conference on Knowledge Discovery and Data Mining (KDD '24), August 25--29, 2024, Barcelona, Spain}\acmDOI{10.1145/3637528.3671977}
\acmISBN{979-8-4007-0490-1/24/08}

\makeatletter
\gdef\@copyrightpermission{
  \begin{minipage}{0.3\columnwidth}
   \href{https://creativecommons.org/licenses/by/4.0/}{\includegraphics[width=0.90\textwidth]{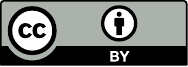}}
  \end{minipage}\hfill
  \begin{minipage}{0.7\columnwidth}
   \href{https://creativecommons.org/licenses/by/4.0/}{This work is licensed under a Creative Commons Attribution International 4.0 License.}
  \end{minipage}
  \vspace{5pt}
}
\makeatother

\usepackage{graphicx}
\usepackage{booktabs}
\usepackage{microtype}
\usepackage{booktabs}
\usepackage{array}
\usepackage{amsfonts}
\usepackage{amsmath}
\usepackage{multirow}
\usepackage{multicol}
\usepackage{listings}
\usepackage{placeins}
\usepackage{subfigure}
\usepackage{graphicx}
\usepackage{enumitem}
\usepackage{hyperref}
\usepackage{tabularx}

\usepackage{xcolor,soul}
\usepackage[most]{tcolorbox}

\newtheorem{problem}{Problem}

\newtheorem{obs}{Observation}

\begin{document}
\newcommand\todo[1]{\textcolor{red}{TODO: #1}}

\def\blue{\textcolor{blue}}

\title{Fake News in Sheep's Clothing: Robust Fake News Detection Against LLM-Empowered Style Attacks}

\author{Jiaying Wu}
\affiliation{
  \institution{National University of Singapore}
  \country{Singapore}
  }
\email{jiayingwu@u.nus.edu}
\author{Jiafeng Guo}
\affiliation{
  \institution{University of Chinese Academy of Sciences}
  \institution{Institute of Computing Technology, CAS}
  \city{Beijing}
  \country{China}
  }
\email{guojiafeng@ict.ac.cn}
\author{Bryan Hooi}
\affiliation{
  \institution{National University of Singapore}
  \country{Singapore}
  }
\email{bhooi@comp.nus.edu.sg}

\begin{abstract}
It is commonly perceived that fake news and real news exhibit distinct writing styles, such as the use of sensationalist versus objective language. However, we emphasize that style-related features can also be exploited for \emph{style-based attacks}. Notably, the advent of powerful Large Language Models (LLMs) has empowered malicious actors to mimic the style of trustworthy news sources, doing so swiftly, cost-effectively, and at scale. Our analysis reveals that LLM-camouflaged fake news content significantly undermines the effectiveness of state-of-the-art text-based detectors (up to 38\% decrease in F1 Score), implying a severe vulnerability to stylistic variations. To address this, we introduce SheepDog, a style-robust fake news detector that prioritizes content over style in determining news veracity. SheepDog achieves this resilience through (1) \emph{LLM-empowered news reframings} that inject style diversity into the training process by customizing articles to match different styles; (2) a \emph{style-agnostic training} scheme that ensures consistent veracity predictions across style-diverse reframings; and (3) \emph{content-focused veracity attributions} that distill content-centric guidelines from LLMs for debunking fake news, offering supplementary cues and potential intepretability that assist veracity prediction. Extensive experiments on three real-world benchmarks demonstrate SheepDog’s style robustness and adaptability to various backbones. 
\footnote{Data and code are available at: \url{https://github.com/jiayingwu19/SheepDog}.}

\end{abstract}

\ccsdesc[500]{Information systems~Data mining}
\ccsdesc[500]{Computing methodologies~Natural language processing}

\keywords{Fake News; Large Language Models; Adversarial Robustness}

\maketitle

\section{Introduction}

\begin{figure}[t]
    \centering
    \includegraphics[width=\columnwidth]{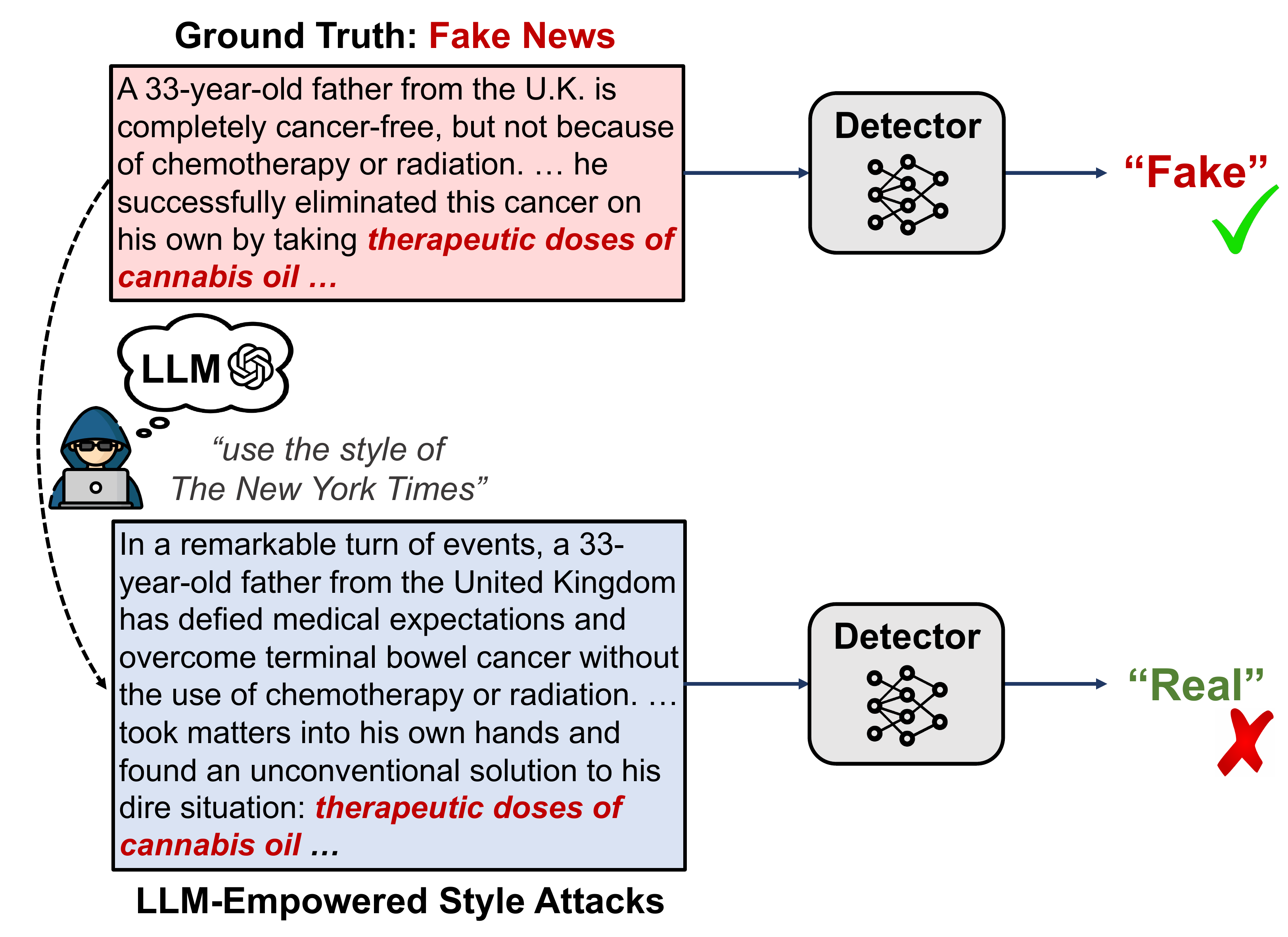}
    \caption{A motivating example of LLM-empowered style attacks on text-based fake news detectors, where fake news is camouflaged with the style of reliable news publishers.}
    \label{fig:sheepdog-intro}
\end{figure}

Psychological theories, such as the Undeutsch hypothesis \cite{amado2015undeutsch}, suggest that genuine and fake statements exhibit distinct linguistic styles. Indeed, reputable news sources uphold journalistic integrity, emphasize accuracy and fact-checking, and maintain a balanced tone \cite{bachmann21defining}. In contrast, unreliable outlets often resort to sensationalism, lack credible sources, and may exhibit partisan biases \cite{higdon21whatis}. Building upon these stylistic differences, recent advances in automated fake news detection have incorporated sentiment features \cite{ajao2019sentiment,zhang2021mining} to enhance the detector, and highlighted the significance of styles in discerning between hyperpartisan news and well-balanced mainstream reporting  \cite{potthast2018stylometric}.

While style-related features serve as key indicators in identifying fake news, they also offer a direct avenue for malicious users to conduct \textbf{style-based attacks}. This problem is exacerbated by the advent of powerful Large Language Models (LLMs) \cite{brown2020language,openai2022chatgpt,ouyang2022training}, whose unprecedented capabilities for reasoning and generative tasks \cite{wei2022emergent,zheng2023chatgpt} bridges the gap between machine-generated and human-written news. Consequently, malicious actors now possess the capability to mimic the style of reputable news sources, in an attempt to evade automated detection. As shown in Figure \ref{fig:sheepdog-intro}, using a style-oriented prompt (i.e., ``style of The New York Times''), LLM-camouflaged fake news successfully bypasses a RoBERTa \cite{liu2019roberta}-based fake news detector.

The impact of stylistic variations on fake news detectors has received limited attention, despite prior investigations into attacks regarding social engagements \cite{wang2023attack}, word-level perturbations \cite{koenders2021vulnerable}, and machine-generated malicious comments \cite{le2020malcom}. To assess the robustness of text-based fake news detectors, we introduce a series of style-based attacks, specifically by tailoring news articles to adversarial writing styles (detailed in Section \ref{sec:attack-formulation}). Our experiments, as summarized in Table \ref{tab:attack-vulnerability}, reveal a significant performance degradation in state-of-the-art text-based detectors, with some suffering up to a 38\% decline in F1 Score. Existing detectors, well-fitted to real news from trustworthy sources and fake news from unreliable sources, struggle to adapt to real-world scenarios where news content is presented in diverse styles. 

To address the \textit{style-related vulnerability} of text-based detectors, we introduce SheepDog, a style-robust approach that consistently recognizes trustworthy content and identifies deceptive content, even when concealed within the LLM-empowered ``sheep's clothing''. Built upon a pretrained language model (LM) backbone, which can be fully fine-tuned to capture task-specific salient features, and leveraging the strong zero-shot reasoning and generative capabilities of LLMs, SheepDog effectively combines the strengths of both (fine-tuned LM: specialized; LLM: versatile).

The style-agnostic nature of SheepDog stems from its utilization of \textit{LLM-empowered news reframings}. To accommodate the wide spectrum of writing styles presented in real-world news articles, we harness the impressive capabilities of LLMs in adhering to complex real-world instructions \cite{he2024large}. By doing so, we can reframe training articles into style-diverse expressions while preserving the integrity of their content. Subsequently, through \textit{style-agnostic training}, we aim to ensure consistency in the model's veracity predictions across each news article and its style-diverse reframings. This training scheme encourages SheepDog to discount style-related features, enabling it to focus on capturing style-agnostic veracity signals from the news content.

To reinforce the emphasis of our approach on news content over style, we propose to incorporate \textit{content-focused veracity attributions} from an LLM to inform veracity predictions. Leveraging the extensive world knowledge and reasoning capabilities within LLMs \cite{asai2024selfrag,menon2023visual}, we elicit explanatory outputs from LLMs regarding the veracity of news articles, with reference to a set of content-centric fake news debunking rationales specifically related to news content (detailed in Section \ref{sec:llm-attributions}). By converting these rationales into precise pseudo-labels, we introduce additional weak supervision that steers SheepDog towards learning robust, style-agnostic news representations. Our approach leverages these attribution-level predictions not only to facilitate style robustness but also to potentially offer explainability into the veracity of news articles. 

Our key contributions are summarized as follows:
\begin{itemize}  [leftmargin=*]
\item \textbf{Empirical Finding}: We present a novel finding on the \textit{style-related vulnerability} of state-of-the-art text-based fake news detectors to LLM-empowered style attacks.
\item \textbf{LLM-Empowered Style Robustness}: We introduce SheepDog, a style-agnostic fake news detector that achieves robustness through style-agnostic training and content-focused veracity attribution prediction, synergized within a multi-task learning paradigm.
\item \textbf{Effectiveness}: Extensive experiments demonstrate that SheepDog achieves significantly superior style robustness across multiple style-based adversarial settings, and yields consistent performance gains when combined with representative LM and LLM backbones.

\end{itemize}

\section{Related Work}

\noindent\textbf{Fake News Detection.} Automated fake news detection has been explored using a wide range of neural architectures \cite{pelrine2021surprising,shu2019defend,zhou2020safe}.
Apart from extracting lexical \cite{rashkin2017truth} and sentiment features \cite{potthast2018stylometric} within the news article text, many methods incorporate auxiliary features to supplement veracity prediction, including user comments \cite{shu2019defend}, news environments \cite{sheng2022zoom}, knowledge bases \cite{cui2020deter,dun2021kan}, temporal patterns from users \cite{ruchansky2017csi}, and social graphs \cite{nguyen2020fang,wu2023decor,wu2023prompt}. Recent studies also seek to address challenges including temporal shift \cite{hu2023learn}, entity bias \cite{zhu2022generalizing} and domain shift \cite{nan2021mdfend,nan2022improving,zhu2022memory} in fake news detection scenarios. In this work, we adopt a text-based perspective, specifically focusing on enhancing the robustness of fake news detectors against stylistic variations.

\noindent\textbf{Adversarial Attack on Fake News Detectors. } Investigating the vulnerabilities of fake news detectors is central to improving their real-world applicability. Hence, existing efforts \cite{he2021petgen,horne2019robust,koenders2021vulnerable,lyu2023interpret,wang2023attack,zhou2019fake} have studied the impact of different attacks from multiple aspects, including manipulation of social engagements \cite{lyu2023interpret,wang2023attack} and user behavior \cite{he2021petgen}, fact distortion \cite{koenders2021vulnerable}, subject-object exchange \cite{zhou2019fake}, and blocking of data availability \cite{horne2019robust}. However, the impact of writing styles remains underexplored. To bridge this gap, we investigate the resilience of text-based detectors against LLM-empowered style attacks, and propose a style-agnostic solution.

\noindent\textbf{LLM Capabilities and Misinformation.} LLMs \cite{openai2022chatgpt,touvron2023llama,openai2023gpt4} have demonstrated remarkable reasoning capabilities that even match or surpass human performance in certain scenarios \cite{wei2022emergent,zheng2023chatgpt}. However, the impressive strengths of LLMs have also attracted increasing attention towards LLM-generated misinformation \cite{kreps2022all}. Recent investigations have found that LLMs can act as high-quality misinformation generators \cite{huang2023faking,lucas2023fighting,pan2023risk,zellers2019defending}, and that LLM-generated misinformation is generally harder to detect \cite{chen2023combating,chen2024can}. On a related front, recent work explore the role of LLMs as fact-checkers \cite{guan2023language,pan2023fact,zhang2023towards} and fake news detectors \cite{chen2024can,pelrine2023reliable}, and leverage the commonsense reasoning capabilities to elicit supplementary explanations from LLMs \cite{asai2024selfrag,he2023harnessing,hu2024bad, menon2023visual} that facilitate a wide range of tasks. Although we also instruct an LLM to generate style-related adversarial articles, our goal is to simulate real-world scenarios where news are presented in diverse styles.  Additionally, instead of leveraging LLMs to make veracity judgments that distinguish false information from the truth \cite{chen2024can,hu2024bad,lucas2023fighting,pelrine2023reliable}, in this work, we investigate the role of LLMs in enhancing the style robustness of text-based fake news detectors, specifically through injecting style-diverse reframings and content-centric cues into the training process.

\section{Problem Definition}
\label{sec:problem-defn}

Let $\mathcal{D}$ be a news dataset consisting of $N$ questionable news pieces, denoted as $p_1,p_2,\dots,p_N$. Among the news pieces, $\mathcal{P}_{L} \subset \mathcal{D}$ is a set of labeled news articles. Each news article in $\mathcal{P}_{L}$ is assigned a ground-truth veracity label $y$. In line with prior work \cite{shu2019defend,zhang2021mining,zhou2020safe}, $y$ is a binary label that represents either real news or fake news.

As we focus on style-related issues, we consider a \textit{text-based} setting. Formally, the problem can be defined as follows:
\begin{problem}
[Text-Based Fake News Detection] Given a news dataset $\mathcal{D}$ with training labels $\mathcal{Y}_{L}$, the goal is to predict the veracity labels of unlabeled news pieces $\mathcal{P}_{U}=\mathcal{D}\setminus\mathcal{P}_{L}$. 
\end{problem}

\section{LLM-Empowered Style Attacks}
\label{sec:style-attacks}
In this section, we establish a series of LLM-empowered style attacks, and conduct preliminary analysis to assess the robustness of state-of-the-art text-based fake news detectors. 

\subsection{Attack Formulation}
\label{sec:attack-formulation}

The impressive capabilities of LLMs \cite{brown2020language,openai2022chatgpt,ouyang2022training} enable malicious users to disguise fake news with restyling prompts, resulting in camouflaged articles that closely resemble reliable sources. In this work, we explore a direct form of style-based attack utilizing \textit{news publisher names} (e.g., ``CNN''). These names possess distinct styles that can be readily adopted by producers of fake news, making them a likely occurrence in real-world scenarios.

To simulate the adversarial situations where news articles are restyled in relation to various publishers, we manipulate the styles of both trustworthy and unreliable news. Specifically, among the test samples, we utlize an LLM to rephrase real news in the style of tabloids, and fake news in the style of mainstream sources. Our general prompt format is shown as follows:
\begin{tcolorbox}[colback=black!2!white,colframe=white!50!black,boxrule=0.5mm]
  Rewrite the following article using the style of [publisher name]: [news article]
\end{tcolorbox}
Based on publisher popularity, in the place of [publisher name], we select ``National Enquirer" to transform real news, and ``CNN'' to transform fake news. These LLM-restyled test articles are then employed to evaluate the resilience of a detector against style-based attacks, as illustrated in Figure \ref{fig:sheepdog-intro}.

\subsection{Style-Related Detector Vulnerability}
\label{sec:observations}
Automated fake news detection becomes increasingly difficult against LLM-empowered style attacks. In this subsection, we conduct preliminary analysis on real-world new articles to evaluate the influence of writing styles on text-based detectors. Our analysis is based on the FakeNewsNet \cite{shu2020fakenewsnet} benchmark (consisting of PolitiFact and GossipCop datasets) and the Labeled Unreliable News (LUN) dataset \cite{rashkin2017truth}, with dataset descriptions relegated to Section \ref{sec:datasets} and Table \ref{tab:ds-stats}. Specifically, we investigate the following question: \textbf{\textit{To what extent can text-based fake news detectors withstand LLM-empowered style attacks?}}
 
In Table \ref{tab:attack-vulnerability}, we examine $13$ representative text-based detectors under both original and adversarial settings (detailed method descriptions are provided in Section \ref{sec:baselines}). These detectors encompass three categories: \textbf{\textit{(1)}} \textit{text-based fake news detectors} with diverse task-specific architectures, including Recurrent Neural Networks (RNNs) \cite{shu2019defend}, Convolutional Neural Networks (CNNs) \cite{zhou2020safe},  Graph Neural Networks (GNNs) applied to document graphs \cite{vaibhav2019sentence}, and Transformers \cite{zhang2021mining};  \textbf{\textit{(2)}} \textit{Fine-tuned LMs} \cite{devlin2019bert,he2021deberta,liu2019roberta,schick2021exploiting,hu2022knowledgeable,xie2020unsupervised} on the fake news detection benchmark datasets; and  \textbf{\textit{(3)}}
\textit{LLMs} \cite{openai2022chatgpt,ouyang2022training,touvron2023llama} with zero-shot prompting. Few-shot and fine-tuned LLM experiments are relegated to Appendix \ref{sec:reframing-llm-robustness}.

We evaluate the robustness of detectors against LLM-empowered style attack based on their performance under the adversarial setting outlined in Section \ref{sec:attack-formulation}. Our empirical results in Table \ref{tab:attack-vulnerability} and Appendix \ref{sec:reframing-llm-robustness} yield the following two implications:

\begin{obs}[Style-Related Vulnerability of Fake News Detectors]
    State-of-the-art text-based fake news detectors are susceptible to LLM-empowered style attacks. This susceptibility results in substantial performance degradation,with an F1 Score decline of up to 38.3\% on the adversarial test set.
\label{obs:vulnerability}
\end{obs}

\begin{obs}[Insufficiency of LLMs as Fake News Detectors]
    LLMs, despite their impressive zero-shot capabilities as general-purpose foundation models, exhibit inferior detection performance compared to text-based fake news detectors and pre-trained LMs fine-tuned specifically for fake news detection.
\label{obs:llm-insufficiency}
\end{obs}

Our two findings suggest a fundamental limitation of text-based fake news detectors in achieving robust veracity predictions against stylistic variations. Detectors overly influenced by styles struggle to reliably differentiate between real and fake news, and even the powerful LLMs may prove inadequate for the specific demands of fake news detection. In the dynamic digital landscape, the styles of news articles evolve rapidly, while the accessibility for malicious users to manipulate style using LLMs exacerbates these variations. Therefore, for effective deployment, a fake news detector must prioritize the assessment of news content over style. This objective motivates our subsequent innovations toward a style-agnostic fake news detection approach.

\begin{table}[t]
\caption{Under LLM-empowered style attacks, existing text-based fake news detectors suffer severe performance deterioration in terms of F1 Score (\%). (O: original; A ($\downarrow$): gap between original unperturbed performance and adversarial performance on the test set formulated in Section \ref{sec:attack-formulation}).}
\centering 
\resizebox{\columnwidth}{!}{
 \begin{tabular}{lcccccccc} \toprule
 \multirow{2}{*}{\textbf{Method}} & \multicolumn{2}{c}{\textbf{PolitiFact}} && \multicolumn{2}{c}{\textbf{GossipCop}} && \multicolumn{2}{c}{\textbf{LUN}}\\
 \cmidrule{2-3} \cmidrule{5-6} \cmidrule{8-9}
 & O & A ($\downarrow$)&& O & A ($\downarrow$) && O & A ($\downarrow$) \\ 
 \toprule
 dEFEND\textbackslash c \cite{shu2019defend}  & 82.59 & 12.15 && 70.74 & 4.34 && 80.92 & 19.16 \\
 SAFE\textbackslash v \cite{zhou2020safe}  & 79.85 & 8.74 && 70.64 & 2.93 && 79.46 & 13.12 \\
 SentGCN \cite{vaibhav2019sentence}  & 80.77 & 13.82 && 69.29 & 5.59 && 79.66 & 16.65 \\
 DualEmo \cite{zhang2021mining}  & 87.76 & 15.34 && 75.36 & 5.89 && 81.52 & 24.97 \\
 \midrule
 BERT \cite{devlin2019bert}  & 84.99 & 12.68 && 74.50 & 5.52 && 80.96 & 24.61 \\
 RoBERTa  \cite{liu2019roberta}  & 87.40 & 11.23 && 74.05 & 3.05 && 82.12 & 29.65 \\
 DeBERTa \cite{he2021deberta}  & 86.30 & 11.73 && 73.80 & 2.85 && 83.67 & 30.34 \\
 UDA \cite{xie2020unsupervised}   & 87.74 & 10.14 && 74.22 & 4.54  && 82.94 & 20.71 \\
 PET \cite{schick2021exploiting}  & 85.51 & 11.02 && 74.63 & 3.08 && 83.66 & 31.08 \\
 KPT  \cite{hu2022knowledgeable}  & 87.70 & 13.26 && 74.23 & 2.63  && 84.06 & 31.83 \\
 \midrule
 GPT-3.5 \cite{openai2022chatgpt}  & 69.61 & 27.48 && 56.30 & 16.71 && 79.97 & 20.34 \\
 InstructGPT \cite{ouyang2022training} & 64.59 & 20.69 && 50.38 & 9.13 && 68.16 &  11.39\\
 LLaMA2-13B \cite{touvron2023llama}  & 63.15 & 29.91 && 53.54 & 27.75 && 70.97 & 38.33\\
 \bottomrule
\end{tabular}
}
\label{tab:attack-vulnerability}
\end{table}

\begin{figure*}[t]
    \centering
    \includegraphics[width=0.95\textwidth]{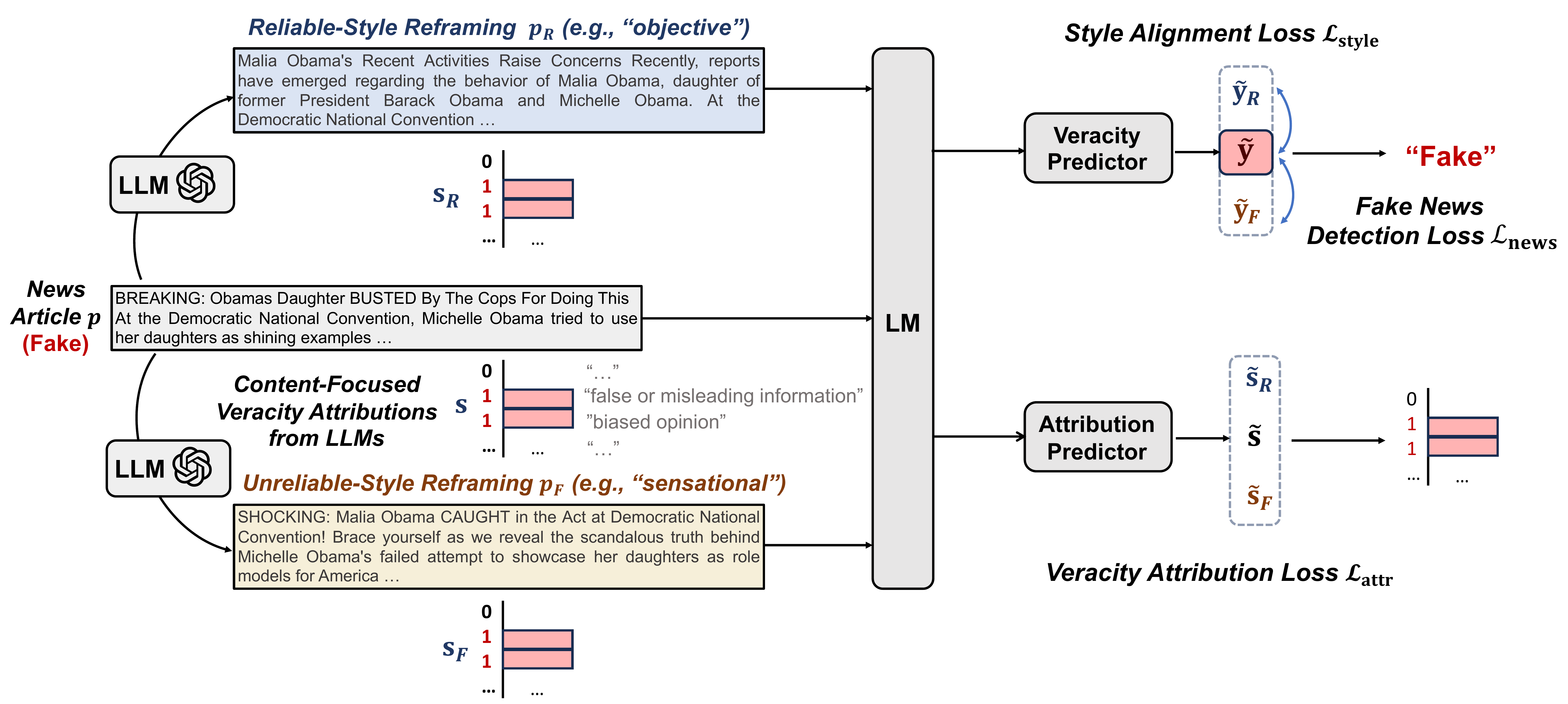}
    \caption{Overview of the proposed SheepDog framework for style-agnostic fake news detection.}
    \label{fig:sheepdog-overview}
\end{figure*}

\section{Proposed Approach}
\label{sec:approach}

Building upon our empirical findings on the style-related vulnerability of text-based fake news detectors, we introduce SheepDog, a style-agnostic detector that reliably assesses news veracity. As overviewed in Figure \ref{fig:sheepdog-overview}, SheepDog obtains style robustness from two core objectives within a multi-task learning paradigm: \textbf{\textit{(1)}} \textbf{style-agnostic training}, which flexibly adapts an LM to the fake news classification task, while ensuring consistent veracity predictions across a diverse array of LLM-empowered news reframings; and \textbf{\textit{(2)}} \textbf{content-focused veracity attribution prediction}, which leverages veracity-related insights from LLMs to inform model predictions. To enhance usability and composability, we design our method to be simple and modular, allowing it to be integrated with any LM and LLM backbone.

\subsection{LLM-Empowered News Reframing}
\label{sec:llm-reframe}

As suggested by our Observation \ref{obs:vulnerability}, text-based fake news detectors fitted to style-consistent real and fake articles exhibit limited adaptability against stylistic variations. To overcome this limitation, our key idea is to inject style diversity into the training stage through a process we term \textit{reframing}, where each news article is presented in various styles.

SheepDog's reframing strategy is driven by two sub-goals:  \textbf{\textit{(1)}} encompassing a wide range of styles and \textbf{\textit{(2)}} maintaining the integrity of the original news content. LLMs, capable of following complex real-world instructions \cite{he2024large}, inherently meet both criteria. This is further validated by our analysis in Appendix \ref{sec:content-consistency}, where LLMs generally prove effective in transforming the tone of news articles while preserving content consistency. Hence, for each training article, we generate a series of prompts, each comprising the news article and a style-oriented reframing instruction.  The general structure of the prompt is as follows, with a detailed example presented in Table \ref{tab:news-reframing-example}:

\begin{tcolorbox}
[colback=black!2!white,colframe=white!50!black,boxrule=0.5mm]

  Rewrite the following article in a / an [specified] tone: [news article]
\end{tcolorbox}
To generate news expressions that simulate both reliable and unreliable sources, we establish a set of four general style-oriented adjectives for the prompt: ``objective and professional'' and ``neutral'' to emulate reliable sources, and ``emotionally triggering'' and ``sensational'' for unreliable sources. During the training stage, for labeled news article $p\in\mathcal{P}_{L}$, we randomly select one reliable-style reframing prompt and one unreliable-style reframing prompt to generate diverse expressions. Through querying the LLM, we obtain two corresponding reframings: one reliable-style reframing denoted as $p_{R}$, and one unreliable-style reframing denoted as $p_{F}$. 

\subsection{Style-Agnostic Training}
\label{sec:pred-consistency}

A style-robust fake news detector must be capable of discerning the veracity of news articles based on their content, without being influenced by stylistic features. To this end, we introduce a \textit{style alignment objective} that ensures close alignment among the veracity predictions of news article $p$, its reliable-style reframing $p_R$, and its unreliable-style reframing $p_F$. This objective is derived as follows.

Let $\mathcal{M}$ be a pre-trained Language Model (LM) such as RoBERTa \cite{liu2019roberta}. Employing an LM as the backbone of the detector offers advantages, as LMs can be readily fine-tuned for the fake news detection task, which enables them to effectively extract salient task-specific features from the news content. Through $\mathcal{M}$, based on $p$, $p_{R}$ and $p_{F}$, we acquire article representations $\mathbf{h} \in \mathbb{R}^{d}$, and reframing representations $\mathbf{h}_R, \mathbf{h}_F\in \mathbb{R}^{d}$:
\begin{equation}
    \mathbf{h}_p = \mathcal{M}(p), \quad \mathbf{h}_{R} = \mathcal{M}(p_{R}), \quad \mathbf{h}_{F} = \mathcal{M}(p_{F}).
\label{eq:news-representations}
\end{equation}

Subsequently, we apply a Multi-Layer Perceptron (MLP) to $p$, $p_{R}$ and $p_{F}$ to obtain corresponding veracity predictions $\tilde{\mathbf{y}}, \tilde{\mathbf{y}}_{R}, \tilde{\mathbf{y}}_{F} \in \mathbb{R}^{2}$: 

\begin{equation}
    \tilde{\mathbf{y}} = \text{MLP}_{pred}(\mathbf{h}), 
    \quad 
    \tilde{\mathbf{y}}_{R} = \text{MLP}_{pred}(\mathbf{h}_{R}), \quad 
    \tilde{\mathbf{y}}_{F} = \text{MLP}_{pred}(\mathbf{h}_{F}).
\label{eq:veracity-pred}
\end{equation}
Each of $\tilde{\mathbf{y}}, \tilde{\mathbf{y}}_{R}, \tilde{\mathbf{y}}_{F}$ contain two logits that correspond to the real and fake classes, respectively.

Despite differences in style, the fundamental news content remains consistent across the original news article $p$ and its reframings $p_{R}$ and $p_{F}$. Ideally, $p_{R}$ and $p_{F}$ should yield the same veracity prediction as $p$. To this end, we formulate the following \textbf{style alignment loss} defined as:
\begin{equation}
    \mathcal{L}_{\text{style}} = \textsf{MEAN}\left( \mathcal{L}_1(\tilde{\mathbf{y}}_R, \tilde{\mathbf{y}}), \; \mathcal{L}_1(\tilde{\mathbf{y}}_F, \tilde{\mathbf{y}})\right),
\label{eq:consistency-loss}
\end{equation}
where $\mathcal{L}_1$ represents the The Kullback-Leibler (KL) divergence loss.

While aligning the predictions of $p_{R}$ and $p_{F}$ with $p$, it is crucial to ensure accurate veracity prediction for $p$. Therefore, we also incorporate a \textbf{fake news detection loss}:
\begin{equation}
    \mathcal{L}_{\text{news}} = \mathcal{L}_2(\tilde{\mathbf{y}}, y),
\label{eq:classification-loss}
\end{equation}
where $\mathcal{L}_2$ represents the standard cross entropy (CE) loss.

\subsection{Content-Focused Veracity Attributions}
\label{sec:llm-attributions}

In addition to tuning the LM backbone with the style alignment objective, which discounts style-related features and encourages style-robust predictions, we further propose to integrate auxiliary veracity-related knowledge and reasoning to inform veracity predictions. To achieve this goal, leveraging the impressive zero-shot reasoning capabilities of general-purpose LLMs \cite{asai2024selfrag,menon2023visual} serves as a promising solution.

Specifically, we elicit \textit{content-focused veracity attributions} from an LLM, which provides explanatory outputs on why each fake news article in the training set is flagged as fake. Our prompt consists of a fake news article and a predefined set $\mathcal{C}$ of content-oriented rationales for debunking fake news (e.g., ``lack of credible sources'' and ``false or misleading information''; detailed rationales are described in Appendix \ref{sec:llm-attribution-prompt}). This prompt efficiently leverages the LLM's reasoning capabilities and prior knowledge to identify characteristics associated with fake news, referencing the rationales in $\mathcal{C}$. The general prompt format is as follows:

\begin{tcolorbox}[colback=black!2!white,colframe=white!50!black,boxrule=0.5mm]
  \textbf{Article}: [fake news article]\\
  \textbf{Question}: [given a list of content-centric rationales for debunking fake news, ask the LLM to identify rationales fulfilled by the fake news article]
\end{tcolorbox}
Upon querying the LLM, we obtain a list of veracity attributions based on the input article. These attributions are then converted into $|\mathcal{C}|$-dimensional pseudo-labels, where each rationale in $\mathcal{C}$ is represented by a distinct binary label. For a given news article $p\in\mathcal{P}_{L}$, this process yields pseudo-labels $\mathbf{s}\in\mathbb{R}^{|\mathcal{C}|}$, which contains supplementary veracity-related information. Similarly, for reframings $p_R$ and $p_F$, we obtain pseudo-labels $\mathbf{s}_{R},\mathbf{s}_{F}\in\mathbb{R}^{|\mathcal{C}|}$, respectively. Notably, since $\mathcal{C}$ focuses solely on fake news indicators, the pseudo-labels for real news and its reframings are uniformly set to all zeros.

To distill the veracity-informed knowledge from these attributions, we introduce a \textit{multi-label attribution prediction} objective.  This enriches our framework with additional content-centric guidance, and offers potential explanability for articles identified as fake during the inference stage (exemplified in Figure \ref{fig:sheepdog-case-study}). 

As shown in Eq. \ref{eq:news-representations}, we learn news representations $\mathbf{h}, \mathbf{h}_R, \mathbf{h}_F\in \mathbb{R}^{d}$ for news article $p$ and its reframings $p_{R}$ and $p_{F}$, respectively, using a pre-trained LM $\mathcal{M}$. Then, the attribution-level prediction scores $\tilde{\mathbf{s}}, \tilde{\mathbf{s}}_{R}, \tilde{\mathbf{s}}_{F} \in \mathbb{R}^{|\mathcal{C}|}$ are computed through another MLP:
\begin{equation}
    \tilde{\mathbf{s}} = \text{MLP}_{attr}(\mathbf{h}), 
    \quad 
    \tilde{\mathbf{s}}_{R} = \text{MLP}_{attr}(\mathbf{h}_{R}), \quad 
    \tilde{\mathbf{s}}_{F} = \text{MLP}_{attr}(\mathbf{h}_{F}).
\label{eq:attr-scores}
\end{equation}

The \textbf{veracity attribution loss} is then defined as:
\begin{equation}
    \mathcal{L}_{\text{attr}} = 
    \textsf{MEAN}\left(
    \mathcal{L}_{3}(\hat{\mathbf{s}}, \; \mathbf{s}),\mathcal{L}_{3}(\hat{\mathbf{s}_R}, \; \mathbf{s}_R),\mathcal{L}_{3}(\hat{\mathbf{s}_F},\; \mathbf{s}_F)\right),
\label{eq:attr-loss}
\end{equation}
where $\mathcal{L}_{3}$ represents the binary cross entropy (BCE) loss, and $\hat{\mathbf{s}}$ denotes the sigmoid-transformed scores in $\tilde{\mathbf{s}}$ corresponding to each rationale.

\subsection{Final Objective Function of SheepDog}
\label{sec:news-pred}

By enforcing consistency among style-diverse news reframings and exploiting the content-focused attributions from the LLM, the final objective function of SheepDog is defined as a linear combination of the the style alignment loss (Eq. \ref{eq:consistency-loss}), the news classification loss (Eq. \ref{eq:classification-loss}), and the veracity attribution loss (Eq. \ref{eq:attr-loss}): 
\begin{equation}
    \mathcal{L} =  \mathcal{L}_{\text{style}} + \mathcal{L}_{\text{news}} +
    \mathcal{L}_{\text{attr}}.
\label{eq:celoss}
\end{equation}

SheepDog is designed as an end-to-end framework, where the style-agnostic news veracity predictor and the content-focused attribution predictor are trained simultaneously.
\begin{table}[t]
\caption{Dataset statistics.}
 \begin{tabular}{lccc} \toprule
 \textbf{Dataset} &  \textbf{PolitiFact} & \textbf{GossipCop} & \textbf{LUN} \\ 
 \toprule
 \# News Articles & 450 & 7,916 & 7,500 \\
 \# Real News & 225 & 3,958 & 3,750 \\
 \# Fake News & 225 & 3,958 & 3,750 \\  
 \bottomrule
\end{tabular} 
 \label{tab:ds-stats}
\end{table}

\begin{table*}[ht]
\caption{SheepDog significantly outperforms competitive baselines on four adversarial test settings under LLM-empowered style attacks (formulated in Section \ref{sec:attack-formulation}), in terms of F1 Score (\%) . Bold (underlined) values indicate the best overall (baseline) performance. Statistical significance over the most competitive baselines, computed using the  Wilcoxon signed-rank test \cite{wilcoxon1945rank}, is indicated with $^{*}$ ($p<.01$). (G1: text-based fake news detectors; G2: LMs fine-tuned to the fake news detection task; G3: LLMs)}
\centering 
\resizebox{\textwidth}{!}{
 \begin{tabular}{clcccccccccccccc} \toprule
\multirow{2}{*}{}& \multirow{2}{*}{\textbf{Method}} & \multicolumn{4}{c}{\textbf{PolitiFact}} && \multicolumn{4}{c}{\textbf{GossipCop}} && \multicolumn{4}{c}{\textbf{LUN}}\\
 \cmidrule{3-6} \cmidrule{8-11} \cmidrule{13-16}
 && A & B & C & D &&  A & B & C & D && A & B & C & D \\ 
 \toprule
  \multirow{4}{*} {\textbf{G1}}& 
 dEFEND\textbackslash c & 70.44 & 69.77 & 73.67 & 72.98 & & 66.40 & 66.55& 68.93 & 69.07 & & 61.76 & 62.28 & 72.95 & 72.50\\
 &SAFE\textbackslash v  & 71.11 & 70.80 & 75.55 & 75.24 & & 67.71 & 67.05& 68.31 & 67.65 & & \underline{66.34} & \underline{67.08} & 72.40 & 73.16\\
 &SentGCN               & 66.95 & 62.50 & 69.54 & 65.08 & & 63.70 & 63.07& 63.61 & 63.01 & & 63.01 & 62.50 & \underline{76.11} & \underline{75.56}\\
 &DualEmo               & 72.42 & 71.23 & 77.07 & 75.80 & & 69.47 & 68.50& 71.69 & 70.71 & & 56.55 & 54.78 & 68.53 & 66.80\\
 \midrule
  \multirow{7}{*} {\textbf{G2}}& 
 BERT     & 72.31 & 71.37 & 77.23 & 76.24 & & 68.98 & 68.17& 71.95 & 71.11 & & 56.35 & 54.61 & 68.50 & 66.74\\
 &RoBERTa & 76.17 & 74.95 & 78.28 & 77.05 & & 71.00 & 70.47& 72.56 & 72.02 & & 52.47 & 53.62 & 68.31 & 69.46\\
 &DeBERTa & 74.57 & 74.36 & \underline{80.60} & \underline{80.35} & & 70.95 & \underline{71.15}& 72.51 & 72.71 & & 53.33 & 55.45 & 67.16 & 69.27\\
 &UDA     & \underline{77.60} & \underline{75.57} & 79.21 & 77.17 & & 69.68 & 69.33& 72.16 & 71.80 & & 62.23 & 61.80 & 68.25 & 67.80\\
 &PET     & 74.49 & 70.75 & 75.49 & 71.76 & & 71.55 & 70.85& \underline{73.74} & 73.02 & & 52.58 & 53.30 & 63.71 & 64.33\\
 &KPT     & 74.44 & 73.32 & 77.73 & 76.60 & & \underline{71.60} & 71.01& 73.69 & \underline{73.10} & & 52.23 & 53.62 & 65.71 & 67.15\\
 \midrule
  \multirow{3}{*} {\textbf{G3}}& 
  GPT-3.5      & 42.13 & 43.44 & 56.61 & 58.17 & & 39.59 & 38.67& 48.44 & 47.38 & & 59.63 & 61.24& 65.74 & 67.43 \\
 &InstructGPT & 43.90 & 43.90 & 54.21 & 54.21 & & 41.25 & 40.18& 44.26 & 43.12 & & 56.77 & 57.15& 58.93 & 59.32 \\
 &LLaMA2-13B  & 33.24 & 34.48 & 53.64 & 55.45 & & 25.79 & 26.06& 37.07 & 37.40 & & 32.64 & 33.00& 50.81 & 51.33 \\
 \midrule
 \textbf{Ours} & SheepDog & \textbf{80.99}$^{*}$ & \textbf{79.89}$^{*}$ & \textbf{82.36}$^{*}$ & \textbf{81.24} & & \textbf{74.45}$^{*}$ & \textbf{74.38}$^{*}$& \textbf{75.95}$^{*}$ & \textbf{75.88}$^{*}$ & & \textbf{85.63}$^{*}$ & \textbf{86.06}$^{*}$& \textbf{87.89}$^{*}$ & \textbf{88.32}$^{*}$ \\
 \bottomrule
\end{tabular} 
}
\label{tab:adv-performance}
\end{table*}

\begin{table}[t]
\caption{Notations and setup for the four style-based adversarial test sets in Section \ref{sec:style-robustness}, denoted as A through D.}
 \begin{tabular}{lcc} \toprule
 \textbf{[publisher name]} &  CNN & The New York Times  \\ 
 \toprule
 National Enquirer & A & B \\
 The Sun & C & D \\ 
 \bottomrule
\end{tabular} 
 \label{tab:adv-setup}
\end{table}

\section{Experiments}
\label{sec:sheepdog-experiments}

In this section, we empirically evaluate SheepDog to investigate the following six research questions:

\begin{itemize}[leftmargin=*]
    \item \textbf{Robustness Against Style Attacks} (Section \ref{sec:style-robustness}): How robust is SheepDog against LLM-empowered style attacks?
    \item \textbf{Effectiveness on Unperturbed Articles} (Section \ref{sec:fnd-performance}): How effectively can SheepDog identify fake news within the original unperturbed test articles?
    \item \textbf{Adaptability to Different Backbones} (Section \ref{sec:lm_backbone}): How well does SheepDog perform when combined with different LM and LLM backbones?
    \item \textbf{Ablation Study} (Section \ref{sec:ablation}): What are the respective roles of style-agnostic training and content-focused attributions on SheepDog's style robustness?
    \item \textbf{Stability Across Reframing Prompts} (Section \ref{sec:prompt-stability}): Does SheepDog yield consistent improvements across diverse sets of news reframing prompts?
    \item \textbf{Case Study} (Section \ref{sec:case_study}): How can we interpret SheepDog's rationale for debunking fake news through its predictions on content-focused veracity attributions?
\end{itemize}

\subsection{Experimental Setup} 
\label{sec:exp-setup}

\subsubsection{Datasets} 
\label{sec:datasets}
We evaluate our approach on three widely-used real-world benchmark datasets: the FakeNewsNet public benchmark \cite{shu2020fakenewsnet}, which consists of the \textbf{PolitiFact} and \textbf{GossipCop} datasets, and the Labeled Unreliable News (\textbf{LUN}) dataset \cite{rashkin2017truth}. Table \ref{tab:ds-stats} describes the dataset statistics. PolitiFact and LUN center on political discourse, while GossipCop focuses on celebrity gossip. For the LUN dataset, which further classifies unreliable news into three sub-categories: satire, hoax, and propaganda, we conduct binary classification between reliable (real) and unreliable (fake) news, and ensure an equal number of unreliable news from each of these fine-grained categories. To better simulate real-world scenarios, we follow prior work \cite{wu2023decor} and adopt temporal data splitting on PolitiFact and GossipCop, where temporal information is available. The most recent 20\% real and fake news articles constitute the test set, and the remaining 80\% articles posted earlier serve as the training set. We adopt random 80/20 training / test splits on LUN.

\subsubsection{Baselines} 
\label{sec:baselines}

We benchmark SheepDog against thirteen representative baseline methods, which can be categorized as:

\textbf{Text-based fake news detectors (G1)} employ neural architectures tailored specifically for the fake news detection task.  \textbf{dEFEND\textbackslash c} is a variant of dEFEND \cite{shu2019defend} based on the news article text that adopts RNN-based hierarchical co-attention. \textbf{SAFE\textbackslash v} is a text-based variant of SAFE \cite{zhou2020safe} that  leverages a CNN-based architecture to learn semantic features. \textbf{SentGCN} \cite{vaibhav2019sentence} encodes veracity-related sentence interaction patterns within each article using a GNN, and
\textbf{DualEmo} \cite{zhang2021mining} incorporates emotion features from news publishers and news comments. As our SheepDog approach does not involve user comments, we implement DualEmo on a BERT-base \cite{devlin2019bert} backbone with publisher emotion features 
for a fair comparison. 

\textbf{Fine-tuned LMs (G2)} adapts pre-trained LMs to the fake news detection task, and has proven effective in handling misinformation scenarios \cite{pelrine2021surprising}.  In addition to three widely-recognized LMs, namely \textbf{BERT} \cite{devlin2019bert}, \textbf{RoBERTa} \cite{liu2019roberta}, and \textbf{DeBERTa} \cite{he2021deberta}, we include \textbf{UDA} \cite{xie2020unsupervised}, a representative BERT-based model that employs diverse text augmentations to yield consistent model predictions against input noise. We also select two methods under the popular \textit{prompting} paradigm: \textbf{PET} \cite{schick2021exploiting}, which converts textual inputs into cloze questions that contain a task description; and \textbf{KPT}, \cite{hu22knowledgeable} which expands the label word space with varied class-related tokens. For a fair comparison, as our proposed approach does not involve unlabeled articles, we implement UDA using consistency training on the supervised training data, and exclude the self-training and PLM ensemble components for PET. All methods in this category are implemented with base version LMs, in line with our approach.

\textbf{LLMs (G3)} conduct zero-shot veracity prediction. We select three representative baseline LLMs: \textbf{GPT-3.5} \cite{openai2022chatgpt}, \textbf{InstructGPT} \cite{ouyang2022training}, and \textbf{LLaMA2-13B} \cite{touvron2023llama}, detailed in
Appendix \ref{sec:llm-detection-prompt}.

\subsubsection{Implementation Details}
We implement SheepDog and its variants based on PyTorch 1.10.0 with CUDA 11.1. We utilize pretrained RoBERTa-base weights from HuggingFace Transformers 4.13.0 \cite{wolf20transformers}. The LM backbone for SheepDog was configured with a maximum sequence length of 512, a batch size of 4, and a learning rate of $2\times10^{-5}$. We prompt GPT-3.5 to generate news reframings, content-focused veracity attributions, and adversarial test articles.  (detailed prompting descriptions and examples are presented in Appendix \ref{sec:llm-reframe-prompt} and \ref{sec:llm-attribution-prompt}). For SheepDog's attribution prediction and veracity prediction, we employ two MLPs, each with a single layer (we also implement a variant with 2-layer MLPs in Section \ref{sec:ablation}). Our model is fine-tuned for $5$ epochs. For the implementation of baseline methods, we adhere to the architectures and hyperparameters recommended by their respective authors. 

We evaluate model performance using Accuracy (\%) and macro-F1 Score (\%). For all experiments except those involving LLMs, we report averaged metrics over 10 runs of each method to provide a comprehensive evaluation. In the case of LLM zero-shot predictions, we employ greedy decoding and conduct each experiment once.

\subsection{Robustness Against Style Attacks}
\label{sec:style-robustness}

We establish a series of LLM-empowered style attacks to assess SheepDog's robustness. Following our prompt template formulated in Section \ref{sec:attack-formulation}, in the place of [publisher name], we select  ``National Enquirer'' and ``The Sun'' to camouflage real news, and ``CNN'' and ``The New York Times'' for fake news, according to publisher popularity. This yields $2\times2=4$ distinct adversarial
test sets, labeled as A through D in Table \ref{tab:adv-setup}. \textbf{\textit{Note that we report the results of SheepDog on adversarial set A in all other subsections, unless otherwise specified.}}

Table \ref{tab:adv-performance} compares the performance of SheepDog with competitive baselines across adversarial test sets A through D under LLM-empowered style attacks. We can observe that: \textit{\textbf{(1)}} All baseline methods are highly susceptible to LLM-empowered style attacks. This vulnerability suggests that existing methods exhibit a tendency towards over-fitting on style-related attributes. \textit{\textbf{(2)}} UDA, which leverages back-translation to generate diverse text augmentations, consistently demonstrates higher robustness compared to its BERT backbone. This suggests the efficacy of incorporating augmentations. However, UDA still struggles to fully adapt to significant stylistic variances in the input articles. This limitation may be attributed to the fact that augmentations through back-translation alone cannot provide sufficient variance. \textit{\textbf{(3)}} On the challenging adversarial test sets of LUN, CNN-based SAFE\textbackslash v and GNN-based SentGCN are more robust than LM-based baselines, which suggests that LMs can be more prone to overfit to style-related features. \textit{\textbf{(4)}} SheepDog outperforms the most competitive baseline by significant margins. Across the three benchmarks, this improvement averages to $2.59$\%, $2.77$\%, and $15.70$\% across the four adversarial test sets, in terms of F1 score. The significantly greater improvements on LUN might be attributed to dataset-specific stylistic attributes, toward which we present a detailed discussion in Section \ref{sec:performance-discussion}.

\begin{table}[t]
\caption{SheepDog achieves performance (\%) that is comparable or superior to competitive baselines on the unperturbed original test sets. Bold (underlined) values indicate the best overall (baseline) performance, and $^{*}$ indicates $p<.01$  using the  Wilcoxon signed-rank test \cite{wilcoxon1945rank}.}
\centering 
\resizebox{\columnwidth}{!}{
 \begin{tabular}{lcccccccc} \toprule
 \multirow{2}{*}{\textbf{Method}} & \multicolumn{2}{c}{\textbf{PolitiFact}} && \multicolumn{2}{c}{\textbf{GossipCop}} && \multicolumn{2}{c}{\textbf{LUN}}\\
 \cmidrule{2-3} \cmidrule{5-6} \cmidrule{8-9}
 & Acc. & F1 && Acc. & F1 && Acc. & F1 \\ 
 \toprule
 dEFEND\textbackslash c  & 82.67 & 82.59 && 70.85 & 70.74  && 81.33 & 80.92 \\
 SAFE\textbackslash v   & 79.89 & 79.85 && 70.71& 70.64 &&  79.93 & 79.46 \\
 SentGCN  & 81.11 & 80.77 && 69.38 & 69.29 && 80.07 & 79.66 \\
 DualEmo  & 87.78 & \underline{87.76} && \underline{75.51} & \underline{75.36} && 81.78 & 81.52 \\
 \midrule
 BERT   & 85.22 & 84.99 && 74.60 & 74.50 && 81.13 & 80.96 \\
 RoBERTa   & \underline{88.00}  & 87.40 && 74.14 & 74.05 && 82.53 & 82.12 \\
 DeBERTa  & 86.33 & 86.30 && 73.86 & 73.80 && 84.01 & 83.67 \\
 UDA    & 87.77 & 87.74 && 74.28 & 74.22 && 83.02 & 82.94 \\
 PET & 85.56 & 85.51 && 74.75 & 74.63 && 84.00 & 83.66 \\
 KPT   & 87.78 & 87.70 && 74.38 & 74.23 && \underline{84.40} & \underline{84.06} \\
 \midrule
 GPT-3.5   & 71.11 & 69.61 && 61.49 & 56.30  && 80.67 & 79.97 \\
 InstructGPT & 67.78 & 64.59 && 58.33 & 50.38 && 70.87 &  68.16
\\
 LLaMA2-13B & 65.56 & 63.15 && 55.74 & 53.54 && 72.47 &  70.97\\
 \midrule
 SheepDog  & \textbf{88.44} & \textbf{88.39} && \textbf{75.77} & \textbf{75.75} && \textbf{93.05}$^{*}$ & \textbf{93.04}$^{*}$ \\

 \bottomrule
\end{tabular}
}
\label{tab:performance}
\end{table}

\begin{table}[t]
\caption{On different LM backbones, SheepDog demonstrates stable and significant improvements (in F1 \%). Statistical significance over the respective LM backbone is computed using the Wilcoxon signed-rank test \cite{wilcoxon1945rank}, denoted by $^{*}$ ($p<.01$).}
{
 \begin{tabular}{lccc} \toprule
 \textbf{Method} &  \textbf{PolitiFact} & \textbf{GossipCop} & \textbf{LUN} \\ 
 \toprule
 RoBERTa  & 76.17 & 71.00 & 52.47 \\
 SheepDog-RoBERTa & \textbf{80.99}$^{*}$ & \textbf{74.45}$^{*}$ & \textbf{85.63}$^{*}$ \\
 \midrule
 BERT  & 72.31  & 68.98 & 53.97 \\
 SheepDog-BERT  & \textbf{81.37}$^{*}$ & \textbf{73.54}$^{*}$ & \textbf{80.36}$^{*}$ \\
 \midrule
 DeBERTa & 74.57  & 70.95 & 53.33 \\
 SheepDog-DeBERTa  & \textbf{81.10}$^{*}$  & \textbf{73.89}$^{*}$ & \textbf{82.58}$^{*}$ \\

 \bottomrule
\end{tabular} 
}
 \label{tab:backbone-lm}
\end{table}

\subsection{Effectiveness on Unperturbed Articles}
\label{sec:fnd-performance}

A desirable fake news detector should achieve style robustness under adversarial settings without compromising its effectiveness under the unperturbed setting. Our empirical results, presented in Table \ref{tab:performance}, demonstrate that SheepDog excels in this regard. When tested on the original, unaltered articles, SheepDog consistently matches (on PolitiFact and GossipCop) or surpasses (on LUN) the performance of the most competitive baseline, in terms of both accuracy and F1 score. Similar to our observation \textit{\textbf{(4)}} in Section \ref{sec:style-robustness}, the significant performance gains on LUN might also stem from dataset-specific stylistic features (detailed in Section \ref{sec:performance-discussion}).

\subsection{Adaptability to LM / LLM Backbones}
\label{sec:lm_backbone}

To assess the flexibility of SheepDog, we evaluate the performance of SheepDog combined with three representative LMs: RoBERTa, BERT and DeBERTa. We also evaluate RoBERTa-based SheepDog combined with two representative LLMs: the closed-source GPT-3.5 and the open-source LLaMA2-13B.

As demonstrated in Table \ref{tab:backbone-lm}, SheepDog \textit{\textbf{(1)}} substantially enhances the performance of each respective LM backbone. Additionally, as shown in Table \ref{tab:adapt-llm}, SheepDog \textit{\textbf{(2)}} also achieves superior style robustness when utilizing both closed-source and open-source LLMs.
This adaptability highlights SheepDog's style robustness from style-invariant training and content-focused attribution prediction, implying its practical utility in real-world scenarios where different LMs and LLMs may be preferred or more readily available.

\begin{table}[t]
\caption{Leveraging closed-source and open-source LLM backbones, SheepDog demonstrates stable and significant improvements (in F1 \%). Statistical significance over the fine-tuned RoBERTa backbone is computed using the Wilcoxon signed-rank test \cite{wilcoxon1945rank}, denoted by $^{*}$ ($p<.01$).}
{
 \begin{tabular}{lccc} \toprule
 \textbf{Method} &  \textbf{PolitiFact} & \textbf{GossipCop} & \textbf{LUN} \\ 
 \toprule
 SheepDog & \textbf{80.99}$^{*}$ & \textbf{74.45}$^{*}$ & \textbf{85.63}$^{*}$ \\
 SheepDog-LLaMA2 & 80.82$^{*}$ & 74.04$^{*}$ & 81.87$^{*}$ \\
 \midrule
 RoBERTa  & 76.17 & 71.00 & 52.47 \\
 \bottomrule
\end{tabular} 
}
 \label{tab:adapt-llm}
\end{table}

\subsection{Ablation Study}
\label{sec:ablation}

To gain deeper insights into the functioning of SheepDog and the role of its different components, we compare SheepDog with the following three model variants:

\begin{itemize}[leftmargin=*]
    \item \textbf{SheepDog w/ 2-layer MLP}, which employs 2-layer MLPs with hidden size of $64$ as attribution detector and veracity detector.
    \item \textbf{SheepDog-R}, which excludes style-diverse news reframings and the style-agnostic training component.
    \item \textbf{SheepDog-A}, which excludes content-focused veracity attributions and the attribution prediction component.
\end{itemize}

\begin{figure}[t]
    \centering
    \includegraphics[width=\columnwidth]{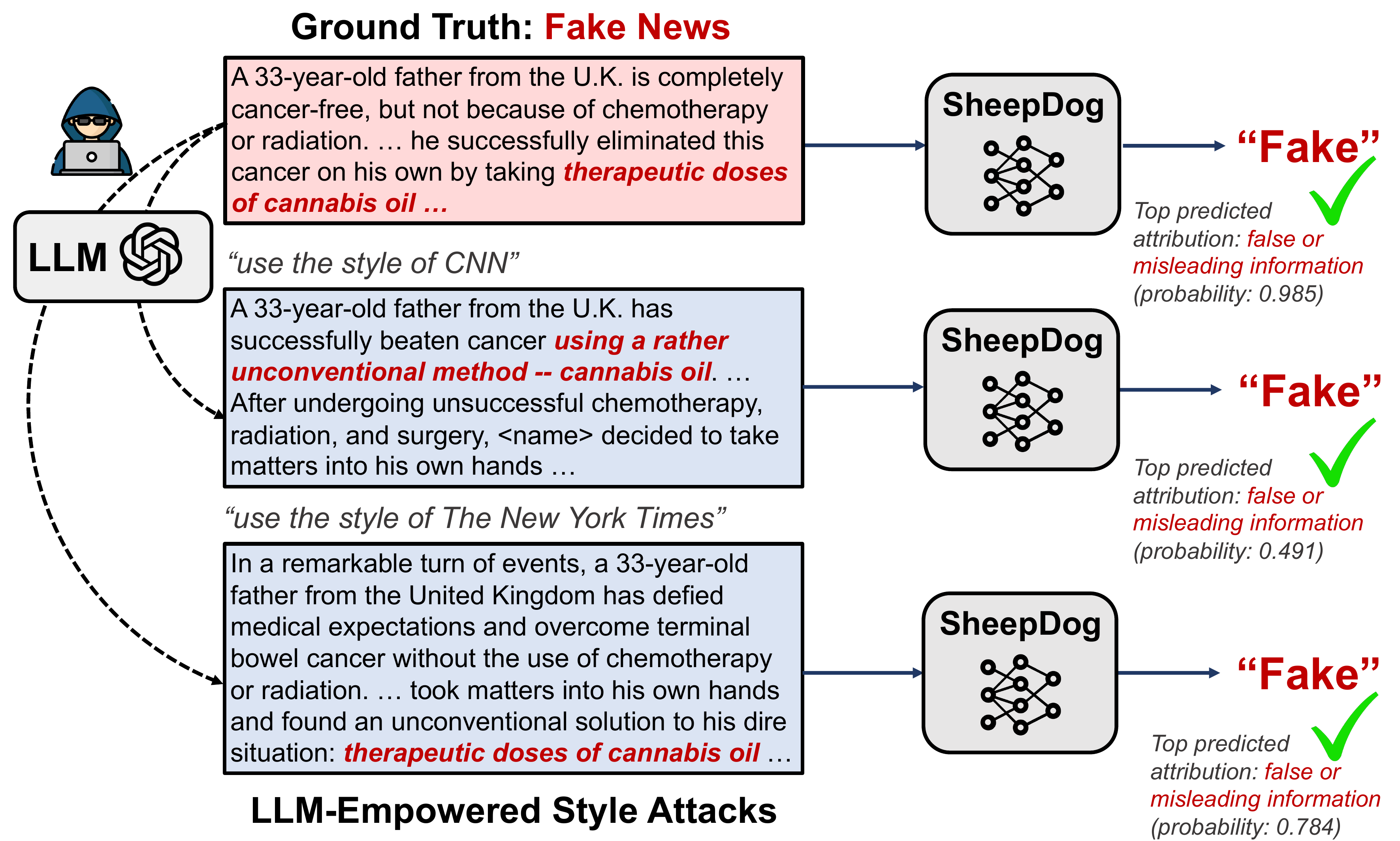}
    \caption{Across the original fake news article and its LLM-camouflaged counterparts, SheepDog maintains consistency and accuracy in both its veracity prediction and the top-predicted veracity attribution for debunking fake news.}
    \label{fig:sheepdog-case-study}
\end{figure}

Results in Table \ref{tab:sheepdog-ablation}  suggest that: \textit{\textbf{(1)}} SheepDog-R without news reframings only yields slight improvements over fine-tuned RoBERTa, which suggests the key role of diverse reframings in SheepDog's robustness. \textit{\textbf{(2)}} While the improvement of SheepDog over SheepDog-A may seem slight, incorporating veracity attributions assists in guiding the model to prioritize content over style. Furthermore, SheepDog's attribution predictor equips the framework with explanatory outputs during inference stage, facilitating easier human verification in real-world scenarios (see Figure \ref{fig:sheepdog-case-study} and Section \ref{sec:case_study} for a concrete illustration of this functionality).
 \textit{\textbf{(3)}} SheepDog, utilizing one single layer for veracity prediction and attribution predictions, slightly outperforms the variant employing 2-layer MLPs. This suggests that the expressiveness of LMs, harnessed through our proposed objectives, yields article representations that contain rich indicators related to both attributions and veracity.

\begin{table}[t]
\caption{Ablation of SheepDog demonstrates benefits of LLM-empowered news reframing (denoted as R) and content-focused veracity attributions (denoted as A) in F1 Score (\%).}
{
 \begin{tabular}{lccc} \toprule
 \textbf{Method} &  \textbf{PolitiFact} & \textbf{GossipCop} & \textbf{LUN} \\ 
 \toprule
 SheepDog  & \textbf{80.99}  & \textbf{74.45} & \textbf{85.63}  \\
 $\;$ w/ 2-layer MLP  & 79.83 & 74.03 & 84.75 \\
 $\;$ - R  & 76.71 & 70.98 & 53.27 \\
 $\;$ - A  & 80.73 & 73.74 & 84.83 \\
 $\;$ RoBERTa  & 76.17 & 71.00& 52.47  \\

 \bottomrule
\end{tabular} 
}
 \label{tab:sheepdog-ablation}
\end{table}

\subsection{Stability Across Reframing Prompts}
\label{sec:prompt-stability}
As detailed in Section \ref{sec:llm-reframe}, we utilize an LLM to generate two types of reframings for a given training article $p$: a reliable-style reframing denoted as $p_R$, and an unreliable-style reframing denoted as $p_F$. To assess SheepDog's stability across different reframing prompts, we examine its performance using four diverse combinations of prompts, denoted as R1 through R4. 

Recall that we adopt the following template for news reframing:

\begin{tcolorbox}[colback=black!2!white,colframe=white!50!black,boxrule=0.5mm]
  Rewrite the following article in a / an [specified] tone: [$p$]
\end{tcolorbox}

For R1 through R4, the specified tones are defined as: ($p_R$ / $p_F$)
\begin{itemize}[leftmargin=*]
    \item \textbf{R1}: ``objective and professional'' /``emotionally triggering''.
    \item \textbf{R2}: ``objective and professional'' / ``sensational''.
    \item \textbf{R3}: ``neutral'' / ``emotionally triggering''.
    \item \textbf{R4}: ``neutral'' / ``sensational''.
\end{itemize}

As shown in Table \ref{tab:prompt-stability}, SheepDog consistently demonstrates stable and significant improvements over the most competitive baseline. This validates the generalizability of our approach, suggesting its potential in effectively combating deceptive information in the ever-evolving digital landscape. Notably, our SheepDog approach conducts sampling between two reliable-style reframings and two unreliable-style reframings, leading to stronger versatility.

\begin{table}[t]
\caption{Across different sets of news reframing prompts, SheepDog demonstrates stable and significant improvements over the most competitive baseline (in F1 \%).}
{
 \begin{tabular}{lccc} \toprule
 \textbf{Method} &  \textbf{PolitiFact} & \textbf{GossipCop} & \textbf{LUN} \\ 
 \toprule
 Baseline (Best)  & 77.60 & 71.60& 66.34  \\
\midrule
 SheepDog  & \textbf{80.99} & \textbf{74.45} & 85.63 \\
 $\;$ w/ R1  & 79.02 & 74.22 & 77.42  \\
 $\;$ w/ R2  & 79.93 & 74.16 &  \textbf{86.18} \\
 $\;$ w/ R3  & 80.36& 73.55 & 76.77\\
 $\;$ w/ R4  & 79.71&74.01 & 85.55  \\

 \bottomrule
\end{tabular} 
}
 \label{tab:prompt-stability}
\end{table}

\subsection{Case Study}
\label{sec:case_study}

To illustrate SheepDog's potential for offering explanatory outputs during model inference via its attribution predictions (detailed in Section \ref{sec:llm-attributions}), we present a case study based on a fake news article from the LUN test set. As shown in Figure \ref{fig:sheepdog-case-study}, the article falsely claims the effectiveness of cannabis oil in treating cancer, aiming to mislead readers, despite contradicting established medical knowledge. While the baseline RoBERTa detector correctly flags the original article as fake news, it misclassifies two style-transformed adversarial articles as real news. In contrast, SheepDog accurately identifies the original article as fake news, a prediction that remains consistent for its two adversarial counterparts. Remarkably, leveraging the softmax-converted probabilities from attribution-level prediction scores (Eq. \ref{eq:attr-scores}), SheepDog consistently identifies "false and misleading information" as the top-predicted attribution for debunking fake news. This style robustness is invaluable for practitioners seeking to comprehend the rationale behind each flagged fake news, aiding in human verification and assessment of prediction reliability.

\subsection{Discussion: Why is SheepDog Yielding Greater Performance Gains on LUN?}
\label{sec:performance-discussion}

SheepDog shows significantly greater performance improvements on LUN compared to PolitiFact and GossipCop in both adversarial (Section \ref{sec:style-robustness}) and original unperturbed settings (Section \ref{sec:fnd-performance}). Specifically, it achieves \textbf{\textit{[A]}} significant improvements on the \textit{original unperturbed} LUN test set while maintaining performance comparable to the best baseline on PolitiFact and GossipCop (Table \ref{tab:performance}); and \textbf{\textit{[B]}} notably greater improvements on LUN compared to PolitiFact and GossipCop on \textit{style-based adversarial} test sets (Table \ref{tab:adv-performance}).

These phenomena can be attributed to the unique style-related features of LUN, which include distinct writing styles of \textbf{\textit{(1)}} individual \textit{news publishers} and and \textbf{\textit{(2)}} different \textit{news types}.  (e.g., hoax). Unlike PolitiFact and GossipCop, LUN’s publishers (i.e., news sites) in the training and test sets do not overlap \cite{rashkin2017truth}, creating an inherent distribution shift between training and test data. Furthermore, as described in Section \ref{sec:datasets}, the fake news articles in LUN encompass satire, hoax, and propaganda with distinctive writing styles. As a result, fake news detectors trained on LUN are expected to be more reliant on writing style.

Recall that our Observation \ref{obs:vulnerability} reveals the heavy reliance of existing text-based detectors on styles rather than news content for veracity prediction. Trained on LUN, the baseline models become overly reliant on styles specific to both publishers and news types, which potentially explains [A] and [B]:
\begin{itemize}[leftmargin=*]
    \item On the \textit{original unperturbed LUN test set}  (Table \ref{tab:performance}), publisher-specific style features used by baseline models fail to generalize to test articles, as test articles are produced by news sites not included in the training data. In contrast, SheepDog, being style-agnostic, remains unaffected by changes in news publisher styles and yields significant improvements.
    \item On the \textit{adversarial LUN test sets} (Table \ref{tab:adv-performance}), both publisher-specific and news type-specific style features utilized by baselines fail to generalize. Notably, LLM-empowered style attacks reverse the styles of both reliable and unreliable news (Section \ref{sec:attack-formulation}). These style variations make type-specific style features detrimental for veracity prediction. In contrast, SheepDog achieves style robustness through LLM-empowered news reframings and content-focused veracity attributions, thereby reliably detecting fake news.
\end{itemize}

\section{Conclusion and Future Work}
In this paper, we address the critical aspect of style-related robustness in fake news detection. Motivated by our empirical finding on the susceptibility of state-of-the-art text-based detectors to LLM-empowered style attacks, we introduce SheepDog, a style-agnostic fake news detector that emphasizes content veracity over style. Jointly leveraging the strengths of task-specific LM backbones and versatile general-purpose LLMs, SheepDog adopts a multi-task learning paradigm, which integrates style-agnostic training and content-focused veracity attribution prediction. Extensive experiments on three real-world benchmarks demonstrate SheepDog's robustness and effectiveness across various style-based adversarial settings, news reframing prompts, and representative backbones. Moving forward, SheepDog lays a solid foundation for developing more resilient and adaptable models in the ever-changing online landscape, and demonstrates promising potential to be further extended to multi-modal scenarios.

\section{Acknowledgements}
This work was supported in part by the National Research Foundation Singapore, NCS Pte. Ltd. and National University of Singapore under the NUS-NCS Joint Laboratory (Grant A-0008542-00-00).

\bibliographystyle{ACM-Reference-Format}
\bibliography{reference}


\begin{thebibliography}{70}


\ifx \showCODEN    \undefined \def \showCODEN     #1{\unskip}     \fi
\ifx \showDOI      \undefined \def \showDOI       #1{#1}\fi
\ifx \showISBNx    \undefined \def \showISBNx     #1{\unskip}     \fi
\ifx \showISBNxiii \undefined \def \showISBNxiii  #1{\unskip}     \fi
\ifx \showISSN     \undefined \def \showISSN      #1{\unskip}     \fi
\ifx \showLCCN     \undefined \def \showLCCN      #1{\unskip}     \fi
\ifx \shownote     \undefined \def \shownote      #1{#1}          \fi
\ifx \showarticletitle \undefined \def \showarticletitle #1{#1}   \fi
\ifx \showURL      \undefined \def \showURL       {\relax}        \fi
\providecommand\bibfield[2]{#2}
\providecommand\bibinfo[2]{#2}
\providecommand\natexlab[1]{#1}
\providecommand\showeprint[2][]{arXiv:#2}

\bibitem[Ajao et~al\mbox{.}(2019)]%
        {ajao2019sentiment}
\bibfield{author}{\bibinfo{person}{Oluwaseun Ajao}, \bibinfo{person}{Deepayan Bhowmik}, {and} \bibinfo{person}{Shahrzad Zargari}.} \bibinfo{year}{2019}\natexlab{}.
\newblock \showarticletitle{Sentiment Aware Fake News Detection on Online Social Networks}. In \bibinfo{booktitle}{\emph{ICASSP}}. \bibinfo{pages}{2507--2511}.
\newblock


\bibitem[Allcott and Gentzkow(2017)]%
        {allcott2017social}
\bibfield{author}{\bibinfo{person}{Hunt Allcott} {and} \bibinfo{person}{Matthew Gentzkow}.} \bibinfo{year}{2017}\natexlab{}.
\newblock \showarticletitle{Social Media and Fake News in the 2016 Election}.
\newblock \bibinfo{journal}{\emph{Journal of Economic Perspectives}} \bibinfo{volume}{31}, \bibinfo{number}{2} (\bibinfo{year}{2017}), \bibinfo{pages}{211--36}.
\newblock


\bibitem[Amado et~al\mbox{.}(2015)]%
        {amado2015undeutsch}
\bibfield{author}{\bibinfo{person}{Bárbara~G. Amado}, \bibinfo{person}{Ramón Arce}, {and} \bibinfo{person}{Francisca Fariña}.} \bibinfo{year}{2015}\natexlab{}.
\newblock \showarticletitle{Undeutsch hypothesis and Criteria Based Content Analysis: A meta-analytic review}.
\newblock \bibinfo{journal}{\emph{The European Journal of Psychology Applied to Legal Context}} \bibinfo{volume}{7}, \bibinfo{number}{1} (\bibinfo{year}{2015}), \bibinfo{pages}{3--12}.
\newblock


\bibitem[Asai et~al\mbox{.}(2024)]%
        {asai2024selfrag}
\bibfield{author}{\bibinfo{person}{Akari Asai}, \bibinfo{person}{Zeqiu Wu}, \bibinfo{person}{Yizhong Wang}, \bibinfo{person}{Avirup Sil}, {and} \bibinfo{person}{Hannaneh Hajishirzi}.} \bibinfo{year}{2024}\natexlab{}.
\newblock \showarticletitle{Self-{RAG}: Learning to Retrieve, Generate, and Critique through Self-Reflection}. In \bibinfo{booktitle}{\emph{ICLR}}.
\newblock


\bibitem[Bachmann et~al\mbox{.}(2021)]%
        {bachmann21defining}
\bibfield{author}{\bibinfo{person}{Philipp Bachmann}, \bibinfo{person}{Mark Eisenegger}, {and} \bibinfo{person}{Diana Ingenhoff}.} \bibinfo{year}{2021}\natexlab{}.
\newblock \showarticletitle{Defining and Measuring News Media Quality: Comparing the Content Perspective and the Audience Perspective}.
\newblock \bibinfo{journal}{\emph{The International Journal of Press/Politics}}  \bibinfo{volume}{27} (\bibinfo{year}{2021}).
\newblock


\bibitem[Bowman et~al\mbox{.}(2015)]%
        {bowman2015large}
\bibfield{author}{\bibinfo{person}{Samuel~R. Bowman}, \bibinfo{person}{Gabor Angeli}, \bibinfo{person}{Christopher Potts}, {and} \bibinfo{person}{Christopher~D. Manning}.} \bibinfo{year}{2015}\natexlab{}.
\newblock \showarticletitle{A large annotated corpus for learning natural language inference}. In \bibinfo{booktitle}{\emph{EMNLP}}. \bibinfo{pages}{632--642}.
\newblock


\bibitem[Brown et~al\mbox{.}(2020)]%
        {brown2020language}
\bibfield{author}{\bibinfo{person}{Tom Brown}, \bibinfo{person}{Benjamin Mann}, \bibinfo{person}{Nick Ryder}, {et~al\mbox{.}}} \bibinfo{year}{2020}\natexlab{}.
\newblock \showarticletitle{Language Models are Few-Shot Learners}. In \bibinfo{booktitle}{\emph{NeurIPS}}, Vol.~\bibinfo{volume}{33}. \bibinfo{pages}{1877--1901}.
\newblock


\bibitem[Chen and Shu(2023)]%
        {chen2023combating}
\bibfield{author}{\bibinfo{person}{Canyu Chen} {and} \bibinfo{person}{Kai Shu}.} \bibinfo{year}{2023}\natexlab{}.
\newblock \showarticletitle{Combating Misinformation in the Age of LLMs: Opportunities and Challenges}.
\newblock \bibinfo{journal}{\emph{arXiv preprint arXiv: 2311.05656}} (\bibinfo{year}{2023}).
\newblock


\bibitem[Chen and Shu(2024)]%
        {chen2024can}
\bibfield{author}{\bibinfo{person}{Canyu Chen} {and} \bibinfo{person}{Kai Shu}.} \bibinfo{year}{2024}\natexlab{}.
\newblock \showarticletitle{Can LLM-Generated Misinformation Be Detected?}. In \bibinfo{booktitle}{\emph{ICLR}}.
\newblock


\bibitem[Cui et~al\mbox{.}(2020)]%
        {cui2020deter}
\bibfield{author}{\bibinfo{person}{Limeng Cui}, \bibinfo{person}{Haeseung Seo}, \bibinfo{person}{Maryam Tabar}, \bibinfo{person}{Fenglong Ma}, \bibinfo{person}{Suhang Wang}, {and} \bibinfo{person}{Dongwon Lee}.} \bibinfo{year}{2020}\natexlab{}.
\newblock \showarticletitle{DETERRENT: Knowledge Guided Graph Attention Network for Detecting Healthcare Misinformation}. In \bibinfo{booktitle}{\emph{KDD}}. \bibinfo{pages}{492–502}.
\newblock


\bibitem[Dettmers et~al\mbox{.}(2023)]%
        {dettmers2023qlora}
\bibfield{author}{\bibinfo{person}{Tim Dettmers}, \bibinfo{person}{Artidoro Pagnoni}, \bibinfo{person}{Ari Holtzman}, {and} \bibinfo{person}{Luke Zettlemoyer}.} \bibinfo{year}{2023}\natexlab{}.
\newblock \showarticletitle{{QL}o{RA}: Efficient Finetuning of Quantized {LLM}s}. In \bibinfo{booktitle}{\emph{NeurIPS}}.
\newblock


\bibitem[Devlin et~al\mbox{.}(2019)]%
        {devlin2019bert}
\bibfield{author}{\bibinfo{person}{Jacob Devlin}, \bibinfo{person}{Ming-Wei Chang}, \bibinfo{person}{Kenton Lee}, {and} \bibinfo{person}{Kristina Toutanova}.} \bibinfo{year}{2019}\natexlab{}.
\newblock \showarticletitle{{BERT}: Pre-training of Deep Bidirectional Transformers for Language Understanding}. In \bibinfo{booktitle}{\emph{NAACL}}. \bibinfo{pages}{4171--4186}.
\newblock


\bibitem[Dun et~al\mbox{.}(2021)]%
        {dun2021kan}
\bibfield{author}{\bibinfo{person}{Yaqian Dun}, \bibinfo{person}{Kefei Tu}, \bibinfo{person}{Chen Chen}, \bibinfo{person}{Chunyan Hou}, {and} \bibinfo{person}{Xiaojie Yuan}.} \bibinfo{year}{2021}\natexlab{}.
\newblock \showarticletitle{KAN: Knowledge-aware Attention Network for Fake News Detection}.
\newblock \bibinfo{journal}{\emph{AAAI}} \bibinfo{volume}{35}, \bibinfo{number}{1} (\bibinfo{year}{2021}), \bibinfo{pages}{81--89}.
\newblock


\bibitem[Guan et~al\mbox{.}(2023)]%
        {guan2023language}
\bibfield{author}{\bibinfo{person}{Jian Guan}, \bibinfo{person}{Jesse Dodge}, \bibinfo{person}{David Wadden}, \bibinfo{person}{Minlie Huang}, {and} \bibinfo{person}{Hao Peng}.} \bibinfo{year}{2023}\natexlab{}.
\newblock \bibinfo{title}{Language Models Hallucinate, but May Excel at Fact Verification}.
\newblock
\newblock
\showeprint[arxiv]{2310.14564}~[cs.CL]


\bibitem[He et~al\mbox{.}(2021a)]%
        {he2021petgen}
\bibfield{author}{\bibinfo{person}{Bing He}, \bibinfo{person}{Mustaque Ahamad}, {and} \bibinfo{person}{Srijan Kumar}.} \bibinfo{year}{2021}\natexlab{a}.
\newblock \showarticletitle{PETGEN: Personalized Text Generation Attack on Deep Sequence Embedding-Based Classification Models}. In \bibinfo{booktitle}{\emph{KDD}}. \bibinfo{pages}{575–584}.
\newblock


\bibitem[He et~al\mbox{.}(2021b)]%
        {he2021deberta}
\bibfield{author}{\bibinfo{person}{Pengcheng He}, \bibinfo{person}{Xiaodong Liu}, \bibinfo{person}{Jianfeng Gao}, {and} \bibinfo{person}{Weizhu Chen}.} \bibinfo{year}{2021}\natexlab{b}.
\newblock \showarticletitle{DeBERTa: Decoding-enhanced BERT with Disentangled Attention}. In \bibinfo{booktitle}{\emph{ICLR}}.
\newblock


\bibitem[He et~al\mbox{.}(2024)]%
        {he2024large}
\bibfield{author}{\bibinfo{person}{Qianyu He}, \bibinfo{person}{Jie Zeng}, \bibinfo{person}{Wenhao Huang}, \bibinfo{person}{Lina Chen}, \bibinfo{person}{Jin Xiao}, \bibinfo{person}{Qianxi He}, \bibinfo{person}{Xunzhe Zhou}, \bibinfo{person}{Lida Chen}, \bibinfo{person}{Xintao Wang}, \bibinfo{person}{Yuncheng Huang}, \bibinfo{person}{Haoning Ye}, \bibinfo{person}{Zihan Li}, \bibinfo{person}{Shisong Chen}, \bibinfo{person}{Yikai Zhang}, \bibinfo{person}{Zhouhong Gu}, \bibinfo{person}{Jiaqing Liang}, {and} \bibinfo{person}{Yanghua Xiao}.} \bibinfo{year}{2024}\natexlab{}.
\newblock \showarticletitle{Can Large Language Models Understand Real-World Complex Instructions?}. In \bibinfo{booktitle}{\emph{AAAI}}. \bibinfo{pages}{18188--18196}.
\newblock


\bibitem[He et~al\mbox{.}(2023)]%
        {he2023harnessing}
\bibfield{author}{\bibinfo{person}{Xiaoxin He}, \bibinfo{person}{Xavier Bresson}, \bibinfo{person}{Thomas Laurent}, \bibinfo{person}{Adam Perold}, \bibinfo{person}{Yann LeCun}, {and} \bibinfo{person}{Bryan Hooi}.} \bibinfo{year}{2023}\natexlab{}.
\newblock \bibinfo{title}{Harnessing Explanations: LLM-to-LM Interpreter for Enhanced Text-Attributed Graph Representation Learning}.
\newblock
\newblock
\showeprint[arxiv]{2305.19523}~[cs.LG]


\bibitem[Higdon(2020)]%
        {higdon21whatis}
\bibfield{author}{\bibinfo{person}{Nolan Higdon}.} \bibinfo{year}{2020}\natexlab{}.
\newblock \showarticletitle{What is Fake News? A Foundational Question for Developing Effective Critical News Literacy Education}.
\newblock \bibinfo{journal}{\emph{Democratic Communiqué}}  \bibinfo{volume}{279} (\bibinfo{year}{2020}).
\newblock
Issue 1.


\bibitem[Horne et~al\mbox{.}(2019)]%
        {horne2019robust}
\bibfield{author}{\bibinfo{person}{Benjamin~D. Horne}, \bibinfo{person}{Jeppe N\o{}rregaard}, {and} \bibinfo{person}{Sibel Adali}.} \bibinfo{year}{2019}\natexlab{}.
\newblock \showarticletitle{Robust Fake News Detection Over Time and Attack}.
\newblock \bibinfo{journal}{\emph{ACM Trans. Intell. Syst. Technol.}} \bibinfo{volume}{11}, \bibinfo{number}{1}, Article \bibinfo{articleno}{7} (\bibinfo{year}{2019}).
\newblock


\bibitem[Hu et~al\mbox{.}(2024)]%
        {hu2024bad}
\bibfield{author}{\bibinfo{person}{Beizhe Hu}, \bibinfo{person}{Qiang Sheng}, \bibinfo{person}{Juan Cao}, \bibinfo{person}{Yuhui Shi}, \bibinfo{person}{Yang Li}, \bibinfo{person}{Danding Wang}, {and} \bibinfo{person}{Peng Qi}.} \bibinfo{year}{2024}\natexlab{}.
\newblock \showarticletitle{Bad Actor, Good Advisor: Exploring the Role of Large Language Models in Fake News Detection}. In \bibinfo{booktitle}{\emph{AAAI}}. \bibinfo{pages}{22105--22113}.
\newblock


\bibitem[Hu et~al\mbox{.}(2023)]%
        {hu2023learn}
\bibfield{author}{\bibinfo{person}{Beizhe Hu}, \bibinfo{person}{Qiang Sheng}, \bibinfo{person}{Juan Cao}, \bibinfo{person}{Yongchun Zhu}, \bibinfo{person}{Danding Wang}, \bibinfo{person}{Zhengjia Wang}, {and} \bibinfo{person}{Zhiwei Jin}.} \bibinfo{year}{2023}\natexlab{}.
\newblock \showarticletitle{Learn over Past, Evolve for Future: Forecasting Temporal Trends for Fake News Detection}. In \bibinfo{booktitle}{\emph{ACL}}. \bibinfo{pages}{116--125}.
\newblock


\bibitem[Hu et~al\mbox{.}(2022a)]%
        {hu2022knowledgeable}
\bibfield{author}{\bibinfo{person}{Shengding Hu}, \bibinfo{person}{Ning Ding}, \bibinfo{person}{Huadong Wang}, \bibinfo{person}{Zhiyuan Liu}, \bibinfo{person}{Jingang Wang}, \bibinfo{person}{Juanzi Li}, \bibinfo{person}{Wei Wu}, {and} \bibinfo{person}{Maosong Sun}.} \bibinfo{year}{2022}\natexlab{a}.
\newblock \showarticletitle{Knowledgeable Prompt-tuning: Incorporating Knowledge into Prompt Verbalizer for Text Classification}. In \bibinfo{booktitle}{\emph{ACL}}. \bibinfo{pages}{2225--2240}.
\newblock


\bibitem[Hu et~al\mbox{.}(2022b)]%
        {hu22knowledgeable}
\bibfield{author}{\bibinfo{person}{Shengding Hu}, \bibinfo{person}{Ning Ding}, \bibinfo{person}{Huadong Wang}, \bibinfo{person}{Zhiyuan Liu}, \bibinfo{person}{Jingang Wang}, \bibinfo{person}{Juanzi Li}, \bibinfo{person}{Wei Wu}, {and} \bibinfo{person}{Maosong Sun}.} \bibinfo{year}{2022}\natexlab{b}.
\newblock \showarticletitle{Knowledgeable Prompt-tuning: Incorporating Knowledge into Prompt Verbalizer for Text Classification}. In \bibinfo{booktitle}{\emph{ACL}}. \bibinfo{pages}{2225--2240}.
\newblock


\bibitem[Huang et~al\mbox{.}(2023)]%
        {huang2023faking}
\bibfield{author}{\bibinfo{person}{Kung-Hsiang Huang}, \bibinfo{person}{Kathleen McKeown}, \bibinfo{person}{Preslav Nakov}, \bibinfo{person}{Yejin Choi}, {and} \bibinfo{person}{Heng Ji}.} \bibinfo{year}{2023}\natexlab{}.
\newblock \showarticletitle{Faking Fake News for Real Fake News Detection: Propaganda-Loaded Training Data Generation}. In \bibinfo{booktitle}{\emph{ACL}}. \bibinfo{pages}{14571--14589}.
\newblock


\bibitem[Jung et~al\mbox{.}(2020)]%
        {jung2020caution}
\bibfield{author}{\bibinfo{person}{Anna-Katharina Jung}, \bibinfo{person}{Bj{\"o}rn Ross}, {and} \bibinfo{person}{Stefan Stieglitz}.} \bibinfo{year}{2020}\natexlab{}.
\newblock \showarticletitle{Caution: Rumors ahead—A case study on the debunking of false information on Twitter}.
\newblock \bibinfo{journal}{\emph{Big Data \& Society}}  \bibinfo{volume}{7} (\bibinfo{year}{2020}).
\newblock


\bibitem[Koenders et~al\mbox{.}(2021)]%
        {koenders2021vulnerable}
\bibfield{author}{\bibinfo{person}{Camille Koenders}, \bibinfo{person}{Johannes Filla}, \bibinfo{person}{Nicolai Schneider}, {and} \bibinfo{person}{Vinicius Woloszyn}.} \bibinfo{year}{2021}\natexlab{}.
\newblock \bibinfo{title}{How Vulnerable Are Automatic Fake News Detection Methods to Adversarial Attacks?}
\newblock
\newblock
\showeprint[arxiv]{2107.07970}~[cs.CL]


\bibitem[Kreps et~al\mbox{.}(2022)]%
        {kreps2022all}
\bibfield{author}{\bibinfo{person}{Sarah Kreps}, \bibinfo{person}{R.~Miles McCain}, {and} \bibinfo{person}{Miles Brundage}.} \bibinfo{year}{2022}\natexlab{}.
\newblock \showarticletitle{All the News That’s Fit to Fabricate: AI-Generated Text as a Tool of Media Misinformation}.
\newblock \bibinfo{journal}{\emph{Journal of Experimental Political Science}} \bibinfo{volume}{9}, \bibinfo{number}{1} (\bibinfo{year}{2022}), \bibinfo{pages}{104–117}.
\newblock


\bibitem[Le et~al\mbox{.}(2020)]%
        {le2020malcom}
\bibfield{author}{\bibinfo{person}{Thai Le}, \bibinfo{person}{Suhang Wang}, {and} \bibinfo{person}{Dongwon Lee}.} \bibinfo{year}{2020}\natexlab{}.
\newblock \showarticletitle{MALCOM: Generating Malicious Comments to Attack Neural Fake News Detection Models}. In \bibinfo{booktitle}{\emph{ICDM}}. \bibinfo{pages}{282--291}.
\newblock


\bibitem[Liu et~al\mbox{.}(2019)]%
        {liu2019roberta}
\bibfield{author}{\bibinfo{person}{Yinhan Liu}, \bibinfo{person}{Myle Ott}, \bibinfo{person}{Naman Goyal}, \bibinfo{person}{Jingfei Du}, \bibinfo{person}{Mandar Joshi}, \bibinfo{person}{Danqi Chen}, \bibinfo{person}{Omer Levy}, \bibinfo{person}{Mike Lewis}, \bibinfo{person}{Luke Zettlemoyer}, {and} \bibinfo{person}{Veselin Stoyanov}.} \bibinfo{year}{2019}\natexlab{}.
\newblock \bibinfo{title}{RoBERTa: A Robustly Optimized BERT Pretraining Approach}.
\newblock
\newblock
\showeprint[arxiv]{1907.11692}~[cs.CL]


\bibitem[Lucas et~al\mbox{.}(2023)]%
        {lucas2023fighting}
\bibfield{author}{\bibinfo{person}{Jason Lucas}, \bibinfo{person}{Adaku Uchendu}, \bibinfo{person}{Michiharu Yamashita}, \bibinfo{person}{Jooyoung Lee}, \bibinfo{person}{Shaurya Rohatgi}, {and} \bibinfo{person}{Dongwon Lee}.} \bibinfo{year}{2023}\natexlab{}.
\newblock \showarticletitle{Fighting Fire with Fire: The Dual Role of LLMs in Crafting and Detecting Elusive Disinformation}. In \bibinfo{booktitle}{\emph{EMNLP}}. \bibinfo{pages}{14279--14305}.
\newblock


\bibitem[Lyu et~al\mbox{.}(2023)]%
        {lyu2023interpret}
\bibfield{author}{\bibinfo{person}{Yuefei Lyu}, \bibinfo{person}{Xiaoyu Yang}, \bibinfo{person}{Jiaxin Liu}, \bibinfo{person}{Sihong Xie}, \bibinfo{person}{Philip Yu}, {and} \bibinfo{person}{Xi Zhang}.} \bibinfo{year}{2023}\natexlab{}.
\newblock \showarticletitle{Interpretable and Effective Reinforcement Learning for Attacking against Graph-based Rumor Detection}. In \bibinfo{booktitle}{\emph{IJCNN}}. \bibinfo{pages}{1--9}.
\newblock


\bibitem[Menon and Vondrick(2023)]%
        {menon2023visual}
\bibfield{author}{\bibinfo{person}{Sachit Menon} {and} \bibinfo{person}{Carl Vondrick}.} \bibinfo{year}{2023}\natexlab{}.
\newblock \showarticletitle{Visual Classification via Description from Large Language Models}. In \bibinfo{booktitle}{\emph{ICLR}}.
\newblock


\bibitem[Nan et~al\mbox{.}(2021)]%
        {nan2021mdfend}
\bibfield{author}{\bibinfo{person}{Qiong Nan}, \bibinfo{person}{Juan Cao}, \bibinfo{person}{Yongchun Zhu}, \bibinfo{person}{Yanyan Wang}, {and} \bibinfo{person}{Jintao Li}.} \bibinfo{year}{2021}\natexlab{}.
\newblock \showarticletitle{MDFEND: Multi-domain fake news detection}. In \bibinfo{booktitle}{\emph{CIKM}}. \bibinfo{pages}{3343--3347}.
\newblock


\bibitem[Nan et~al\mbox{.}(2022)]%
        {nan2022improving}
\bibfield{author}{\bibinfo{person}{Qiong Nan}, \bibinfo{person}{Danding Wang}, \bibinfo{person}{Yongchun Zhu}, \bibinfo{person}{Qiang Sheng}, \bibinfo{person}{Yuhui Shi}, \bibinfo{person}{Juan Cao}, {and} \bibinfo{person}{Jintao Li}.} \bibinfo{year}{2022}\natexlab{}.
\newblock \showarticletitle{Improving Fake News Detection of Influential Domain via Domain- and Instance-Level Transfer}. In \bibinfo{booktitle}{\emph{COLING}}. \bibinfo{pages}{2834--2848}.
\newblock


\bibitem[Nguyen et~al\mbox{.}(2020)]%
        {nguyen2020fang}
\bibfield{author}{\bibinfo{person}{Van-Hoang Nguyen}, \bibinfo{person}{Kazunari Sugiyama}, \bibinfo{person}{Preslav Nakov}, {and} \bibinfo{person}{Min-Yen Kan}.} \bibinfo{year}{2020}\natexlab{}.
\newblock \showarticletitle{FANG: Leveraging Social Context for Fake News Detection Using Graph Representation}. In \bibinfo{booktitle}{\emph{CIKM}}. \bibinfo{pages}{1165–1174}.
\newblock


\bibitem[OpenAI(2022)]%
        {openai2022chatgpt}
\bibfield{author}{\bibinfo{person}{OpenAI}.} \bibinfo{year}{2022}\natexlab{}.
\newblock \bibinfo{title}{{ChatGPT}: Optimizing language models for dialogue}.
\newblock
\newblock


\bibitem[OpenAI(2023)]%
        {openai2023gpt4}
\bibfield{author}{\bibinfo{person}{OpenAI}.} \bibinfo{year}{2023}\natexlab{}.
\newblock \bibinfo{title}{{GPT-4} Technical Report}.
\newblock
\newblock
\showeprint[arxiv]{2303.08774}~[cs.CL]


\bibitem[Ouyang et~al\mbox{.}(2022)]%
        {ouyang2022training}
\bibfield{author}{\bibinfo{person}{Long Ouyang}, \bibinfo{person}{Jeffrey Wu}, \bibinfo{person}{Xu Jiang}, {et~al\mbox{.}}} \bibinfo{year}{2022}\natexlab{}.
\newblock \showarticletitle{Training language models to follow instructions with human feedback}. In \bibinfo{booktitle}{\emph{NeurIPS}}, Vol.~\bibinfo{volume}{35}. \bibinfo{pages}{27730--27744}.
\newblock


\bibitem[Pan et~al\mbox{.}(2023b)]%
        {pan2023fact}
\bibfield{author}{\bibinfo{person}{Liangming Pan}, \bibinfo{person}{Xiaobao Wu}, \bibinfo{person}{Xinyuan Lu}, \bibinfo{person}{Anh~Tuan Luu}, \bibinfo{person}{William~Yang Wang}, \bibinfo{person}{Min-Yen Kan}, {and} \bibinfo{person}{Preslav Nakov}.} \bibinfo{year}{2023}\natexlab{b}.
\newblock \showarticletitle{Fact-Checking Complex Claims with Program-Guided Reasoning}.
\newblock  (\bibinfo{year}{2023}), \bibinfo{pages}{6981--7004}.
\newblock


\bibitem[Pan et~al\mbox{.}(2023a)]%
        {pan2023risk}
\bibfield{author}{\bibinfo{person}{Yikang Pan}, \bibinfo{person}{Liangming Pan}, \bibinfo{person}{Wenhu Chen}, \bibinfo{person}{Preslav Nakov}, \bibinfo{person}{Min-Yen Kan}, {and} \bibinfo{person}{William Wang}.} \bibinfo{year}{2023}\natexlab{a}.
\newblock \showarticletitle{On the Risk of Misinformation Pollution with Large Language Models}. In \bibinfo{booktitle}{\emph{Findings of EMNLP 2023}}. \bibinfo{pages}{1389--1403}.
\newblock


\bibitem[Pelrine et~al\mbox{.}(2021)]%
        {pelrine2021surprising}
\bibfield{author}{\bibinfo{person}{Kellin Pelrine}, \bibinfo{person}{Jacob Danovitch}, {and} \bibinfo{person}{Reihaneh Rabbany}.} \bibinfo{year}{2021}\natexlab{}.
\newblock \showarticletitle{The Surprising Performance of Simple Baselines for Misinformation Detection}. In \bibinfo{booktitle}{\emph{WWW}}. \bibinfo{pages}{3432–3441}.
\newblock


\bibitem[Pelrine et~al\mbox{.}(2023)]%
        {pelrine2023reliable}
\bibfield{author}{\bibinfo{person}{Kellin Pelrine}, \bibinfo{person}{Anne Imouza}, \bibinfo{person}{Camille Thibault}, \bibinfo{person}{Meilina Reksoprodjo}, \bibinfo{person}{Caleb Gupta}, \bibinfo{person}{Joel Christoph}, \bibinfo{person}{Jean-François Godbout}, {and} \bibinfo{person}{Reihaneh Rabbany}.} \bibinfo{year}{2023}\natexlab{}.
\newblock \bibinfo{title}{Towards Reliable Misinformation Mitigation: Generalization, Uncertainty, and GPT-4}.
\newblock
\newblock
\showeprint[arxiv]{2305.14928}~[cs.CL]


\bibitem[Potthast et~al\mbox{.}(2018)]%
        {potthast2018stylometric}
\bibfield{author}{\bibinfo{person}{Martin Potthast}, \bibinfo{person}{Johannes Kiesel}, \bibinfo{person}{Kevin Reinartz}, \bibinfo{person}{Janek Bevendorff}, {and} \bibinfo{person}{Benno Stein}.} \bibinfo{year}{2018}\natexlab{}.
\newblock \showarticletitle{A Stylometric Inquiry into Hyperpartisan and Fake News}. In \bibinfo{booktitle}{\emph{ACL}}. \bibinfo{pages}{231--240}.
\newblock


\bibitem[Rashkin et~al\mbox{.}(2017)]%
        {rashkin2017truth}
\bibfield{author}{\bibinfo{person}{Hannah Rashkin}, \bibinfo{person}{Eunsol Choi}, \bibinfo{person}{Jin~Yea Jang}, \bibinfo{person}{Svitlana Volkova}, {and} \bibinfo{person}{Yejin Choi}.} \bibinfo{year}{2017}\natexlab{}.
\newblock \showarticletitle{Truth of Varying Shades: Analyzing Language in Fake News and Political Fact-Checking}. In \bibinfo{booktitle}{\emph{EMNLP}}. \bibinfo{pages}{2931--2937}.
\newblock


\bibitem[Ruchansky et~al\mbox{.}(2017)]%
        {ruchansky2017csi}
\bibfield{author}{\bibinfo{person}{Natali Ruchansky}, \bibinfo{person}{Sungyong Seo}, {and} \bibinfo{person}{Yan Liu}.} \bibinfo{year}{2017}\natexlab{}.
\newblock \showarticletitle{CSI: A Hybrid Deep Model for Fake News Detection}. In \bibinfo{booktitle}{\emph{CIKM}}. \bibinfo{pages}{797–806}.
\newblock


\bibitem[Schick and Sch{\"u}tze(2021)]%
        {schick2021exploiting}
\bibfield{author}{\bibinfo{person}{Timo Schick} {and} \bibinfo{person}{Hinrich Sch{\"u}tze}.} \bibinfo{year}{2021}\natexlab{}.
\newblock \showarticletitle{Exploiting Cloze-Questions for Few-Shot Text Classification and Natural Language Inference}. In \bibinfo{booktitle}{\emph{EACL}}. \bibinfo{pages}{255--269}.
\newblock


\bibitem[Sheng et~al\mbox{.}(2022)]%
        {sheng2022zoom}
\bibfield{author}{\bibinfo{person}{Qiang Sheng}, \bibinfo{person}{Juan Cao}, \bibinfo{person}{Xueyao Zhang}, \bibinfo{person}{Rundong Li}, \bibinfo{person}{Danding Wang}, {and} \bibinfo{person}{Yongchun Zhu}.} \bibinfo{year}{2022}\natexlab{}.
\newblock \showarticletitle{Zoom Out and Observe: News Environment Perception for Fake News Detection}. In \bibinfo{booktitle}{\emph{ACL}}. \bibinfo{pages}{4543--4556}.
\newblock


\bibitem[Shu et~al\mbox{.}(2019)]%
        {shu2019defend}
\bibfield{author}{\bibinfo{person}{Kai Shu}, \bibinfo{person}{Limeng Cui}, \bibinfo{person}{Suhang Wang}, \bibinfo{person}{Dongwon Lee}, {and} \bibinfo{person}{Huan Liu}.} \bibinfo{year}{2019}\natexlab{}.
\newblock \showarticletitle{DEFEND: Explainable Fake News Detection}. In \bibinfo{booktitle}{\emph{KDD}}. \bibinfo{pages}{395–405}.
\newblock


\bibitem[Shu et~al\mbox{.}(2020)]%
        {shu2020fakenewsnet}
\bibfield{author}{\bibinfo{person}{Kai Shu}, \bibinfo{person}{Deepak Mahudeswaran}, \bibinfo{person}{Suhang Wang}, \bibinfo{person}{Dongwon Lee}, {and} \bibinfo{person}{Huan Liu}.} \bibinfo{year}{2020}\natexlab{}.
\newblock \showarticletitle{FakeNewsNet: A Data Repository with News Content, Social Context, and Spatiotemporal Information for Studying Fake News on Social Media}.
\newblock \bibinfo{journal}{\emph{Big Data}} \bibinfo{volume}{8}, \bibinfo{number}{3} (\bibinfo{year}{2020}), \bibinfo{pages}{171--188}.
\newblock


\bibitem[Shu et~al\mbox{.}(2017)]%
        {shu2017fake}
\bibfield{author}{\bibinfo{person}{Kai Shu}, \bibinfo{person}{Amy Sliva}, \bibinfo{person}{Suhang Wang}, \bibinfo{person}{Jiliang Tang}, {and} \bibinfo{person}{Huan Liu}.} \bibinfo{year}{2017}\natexlab{}.
\newblock \showarticletitle{Fake News Detection on Social Media: A Data Mining Perspective}.
\newblock \bibinfo{journal}{\emph{SIGKDD Explor. Newsl.}} \bibinfo{volume}{19}, \bibinfo{number}{1} (\bibinfo{year}{2017}), \bibinfo{pages}{22–36}.
\newblock


\bibitem[Touvron et~al\mbox{.}(2023)]%
        {touvron2023llama}
\bibfield{author}{\bibinfo{person}{Hugo Touvron}, \bibinfo{person}{Louis Martin}, \bibinfo{person}{Kevin Stone}, {et~al\mbox{.}}} \bibinfo{year}{2023}\natexlab{}.
\newblock \bibinfo{title}{Llama 2: Open Foundation and Fine-Tuned Chat Models}.
\newblock
\newblock
\showeprint[arxiv]{2307.09288}~[cs.CL]


\bibitem[Vaibhav et~al\mbox{.}(2019)]%
        {vaibhav2019sentence}
\bibfield{author}{\bibinfo{person}{Vaibhav Vaibhav}, \bibinfo{person}{Raghuram Mandyam}, {and} \bibinfo{person}{Eduard Hovy}.} \bibinfo{year}{2019}\natexlab{}.
\newblock \showarticletitle{Do Sentence Interactions Matter? Leveraging Sentence Level Representations for Fake News Classification}. In \bibinfo{booktitle}{\emph{TextGraphs-13}}. \bibinfo{pages}{134--139}.
\newblock


\bibitem[Wang et~al\mbox{.}(2023)]%
        {wang2023attack}
\bibfield{author}{\bibinfo{person}{Haoran Wang}, \bibinfo{person}{Yingtong Dou}, \bibinfo{person}{Canyu Chen}, \bibinfo{person}{Lichao Sun}, \bibinfo{person}{Philip~S. Yu}, {and} \bibinfo{person}{Kai Shu}.} \bibinfo{year}{2023}\natexlab{}.
\newblock \showarticletitle{Attacking Fake News Detectors via Manipulating News Social Engagement}. In \bibinfo{booktitle}{\emph{WWW}}. \bibinfo{pages}{3978–3986}.
\newblock


\bibitem[Wei et~al\mbox{.}(2022)]%
        {wei2022emergent}
\bibfield{author}{\bibinfo{person}{Jason Wei}, \bibinfo{person}{Yi Tay}, \bibinfo{person}{Rishi Bommasani}, \bibinfo{person}{Colin Raffel}, \bibinfo{person}{Barret Zoph}, \bibinfo{person}{Sebastian Borgeaud}, \bibinfo{person}{Dani Yogatama}, \bibinfo{person}{Maarten Bosma}, \bibinfo{person}{Denny Zhou}, \bibinfo{person}{Donald Metzler}, \bibinfo{person}{Ed~H. Chi}, \bibinfo{person}{Tatsunori Hashimoto}, \bibinfo{person}{Oriol Vinyals}, \bibinfo{person}{Percy Liang}, \bibinfo{person}{Jeff Dean}, {and} \bibinfo{person}{William Fedus}.} \bibinfo{year}{2022}\natexlab{}.
\newblock \showarticletitle{Emergent Abilities of Large Language Models}.
\newblock \bibinfo{journal}{\emph{Transactions on Machine Learning Research}} (\bibinfo{year}{2022}).
\newblock


\bibitem[Wilcoxon(1945)]%
        {wilcoxon1945rank}
\bibfield{author}{\bibinfo{person}{Frank Wilcoxon}.} \bibinfo{year}{1945}\natexlab{}.
\newblock \showarticletitle{Individual Comparisons by Ranking Methods}.
\newblock \bibinfo{journal}{\emph{Biometrics Bulletin}}  \bibinfo{volume}{1} (\bibinfo{year}{1945}), \bibinfo{pages}{80--83}.
\newblock


\bibitem[Williams et~al\mbox{.}(2018)]%
        {williams2018mnli}
\bibfield{author}{\bibinfo{person}{Adina Williams}, \bibinfo{person}{Nikita Nangia}, {and} \bibinfo{person}{Samuel Bowman}.} \bibinfo{year}{2018}\natexlab{}.
\newblock \showarticletitle{A Broad-Coverage Challenge Corpus for Sentence Understanding through Inference}. In \bibinfo{booktitle}{\emph{NAACL}}. \bibinfo{pages}{1112--1122}.
\newblock


\bibitem[Wolf et~al\mbox{.}(2020)]%
        {wolf20transformers}
\bibfield{author}{\bibinfo{person}{Thomas Wolf}, \bibinfo{person}{Lysandre Debut}, \bibinfo{person}{Victor Sanh}, \bibinfo{person}{Julien Chaumond}, \bibinfo{person}{Clement Delangue}, \bibinfo{person}{Anthony Moi}, \bibinfo{person}{Pierric Cistac}, \bibinfo{person}{Tim Rault}, \bibinfo{person}{Rémi Louf}, \bibinfo{person}{Morgan Funtowicz}, {et~al\mbox{.}}} \bibinfo{year}{2020}\natexlab{}.
\newblock \showarticletitle{Transformers: State-of-the-Art Natural Language Processing}. In \bibinfo{booktitle}{\emph{EMNLP}}. \bibinfo{pages}{38--45}.
\newblock


\bibitem[Wu and Hooi(2023)]%
        {wu2023decor}
\bibfield{author}{\bibinfo{person}{Jiaying Wu} {and} \bibinfo{person}{Bryan Hooi}.} \bibinfo{year}{2023}\natexlab{}.
\newblock \showarticletitle{DECOR: Degree-Corrected Social Graph Refinement for Fake News Detection}. In \bibinfo{booktitle}{\emph{KDD}}. \bibinfo{pages}{2582–2593}.
\newblock


\bibitem[Wu et~al\mbox{.}(2023)]%
        {wu2023prompt}
\bibfield{author}{\bibinfo{person}{Jiaying Wu}, \bibinfo{person}{Shen Li}, \bibinfo{person}{Ailin Deng}, \bibinfo{person}{Miao Xiong}, {and} \bibinfo{person}{Bryan Hooi}.} \bibinfo{year}{2023}\natexlab{}.
\newblock \showarticletitle{Prompt-and-Align: Prompt-Based Social Alignment for Few-Shot Fake News Detection}. In \bibinfo{booktitle}{\emph{CIKM}}. \bibinfo{pages}{2726–2736}.
\newblock


\bibitem[Xie et~al\mbox{.}(2020)]%
        {xie2020unsupervised}
\bibfield{author}{\bibinfo{person}{Qizhe Xie}, \bibinfo{person}{Zihang Dai}, \bibinfo{person}{Eduard Hovy}, \bibinfo{person}{Thang Luong}, {and} \bibinfo{person}{Quoc Le}.} \bibinfo{year}{2020}\natexlab{}.
\newblock \showarticletitle{Unsupervised Data Augmentation for Consistency Training}. In \bibinfo{booktitle}{\emph{NeurIPS}}, Vol.~\bibinfo{volume}{33}. \bibinfo{pages}{6256--6268}.
\newblock


\bibitem[Zellers et~al\mbox{.}(2019)]%
        {zellers2019defending}
\bibfield{author}{\bibinfo{person}{Rowan Zellers}, \bibinfo{person}{Ari Holtzman}, \bibinfo{person}{Hannah Rashkin}, \bibinfo{person}{Yonatan Bisk}, \bibinfo{person}{Ali Farhadi}, \bibinfo{person}{Franziska Roesner}, {and} \bibinfo{person}{Yejin Choi}.} \bibinfo{year}{2019}\natexlab{}.
\newblock \showarticletitle{Defending Against Neural Fake News}. In \bibinfo{booktitle}{\emph{NeurIPS}}, Vol.~\bibinfo{volume}{32}.
\newblock


\bibitem[Zhang et~al\mbox{.}(2018)]%
        {zhang2018structured}
\bibfield{author}{\bibinfo{person}{Amy~X. Zhang}, \bibinfo{person}{Aditya Ranganathan}, \bibinfo{person}{Sarah~Emlen Metz}, \bibinfo{person}{Scott Appling}, \bibinfo{person}{Connie~Moon Sehat}, \bibinfo{person}{Norman Gilmore}, \bibinfo{person}{Nick~B. Adams}, \bibinfo{person}{Emmanuel Vincent}, \bibinfo{person}{Jennifer Lee}, \bibinfo{person}{Martin Robbins}, \bibinfo{person}{Ed Bice}, \bibinfo{person}{Sandro Hawke}, \bibinfo{person}{David Karger}, {and} \bibinfo{person}{An~Xiao Mina}.} \bibinfo{year}{2018}\natexlab{}.
\newblock \showarticletitle{A Structured Response to Misinformation: Defining and Annotating Credibility Indicators in News Articles}. In \bibinfo{booktitle}{\emph{Companion Proceedings of WWW}}. \bibinfo{pages}{603–612}.
\newblock


\bibitem[Zhang et~al\mbox{.}(2021)]%
        {zhang2021mining}
\bibfield{author}{\bibinfo{person}{Xueyao Zhang}, \bibinfo{person}{Juan Cao}, \bibinfo{person}{Xirong Li}, \bibinfo{person}{Qiang Sheng}, \bibinfo{person}{Lei Zhong}, {and} \bibinfo{person}{Kai Shu}.} \bibinfo{year}{2021}\natexlab{}.
\newblock \showarticletitle{Mining Dual Emotion for Fake News Detection}. In \bibinfo{booktitle}{\emph{WWW}}. \bibinfo{pages}{3465–3476}.
\newblock


\bibitem[Zhang and Gao(2023)]%
        {zhang2023towards}
\bibfield{author}{\bibinfo{person}{Xuan Zhang} {and} \bibinfo{person}{Wei Gao}.} \bibinfo{year}{2023}\natexlab{}.
\newblock \bibinfo{title}{Towards LLM-based Fact Verification on News Claims with a Hierarchical Step-by-Step Prompting Method}.
\newblock
\newblock
\showeprint[arxiv]{2310.00305}~[cs.CL]


\bibitem[Zheng and Zhan(2023)]%
        {zheng2023chatgpt}
\bibfield{author}{\bibinfo{person}{Haoyi Zheng} {and} \bibinfo{person}{Huichun Zhan}.} \bibinfo{year}{2023}\natexlab{}.
\newblock \showarticletitle{ChatGPT in Scientific Writing: A Cautionary Tale}.
\newblock \bibinfo{journal}{\emph{The American Journal of Medicine}} \bibinfo{volume}{136}, \bibinfo{number}{8} (\bibinfo{year}{2023}), \bibinfo{pages}{725--726.e6}.
\newblock


\bibitem[Zhou et~al\mbox{.}(2020)]%
        {zhou2020safe}
\bibfield{author}{\bibinfo{person}{Xinyi Zhou}, \bibinfo{person}{Jindi Wu}, {and} \bibinfo{person}{Reza Zafarani}.} \bibinfo{year}{2020}\natexlab{}.
\newblock \showarticletitle{SAFE: Similarity-Aware Multi-modal Fake News Detection}. In \bibinfo{booktitle}{\emph{PAKDD}}. \bibinfo{pages}{354--367}.
\newblock


\bibitem[Zhou et~al\mbox{.}(2019)]%
        {zhou2019fake}
\bibfield{author}{\bibinfo{person}{Zhixuan Zhou}, \bibinfo{person}{Huankang Guan}, \bibinfo{person}{Meghana Bhat}, {and} \bibinfo{person}{Justin Hsu}.} \bibinfo{year}{2019}\natexlab{}.
\newblock \showarticletitle{Fake News Detection via {NLP} is Vulnerable to Adversarial Attacks}. In \bibinfo{booktitle}{\emph{Proceedings of the 11th International Conference on Agents and Artificial Intelligence}}.
\newblock


\bibitem[Zhu et~al\mbox{.}(2022a)]%
        {zhu2022generalizing}
\bibfield{author}{\bibinfo{person}{Yongchun Zhu}, \bibinfo{person}{Qiang Sheng}, \bibinfo{person}{Juan Cao}, \bibinfo{person}{Shuokai Li}, \bibinfo{person}{Danding Wang}, {and} \bibinfo{person}{Fuzhen Zhuang}.} \bibinfo{year}{2022}\natexlab{a}.
\newblock \showarticletitle{Generalizing to the Future: Mitigating Entity Bias in Fake News Detection}. In \bibinfo{booktitle}{\emph{SIGIR}}.
\newblock


\bibitem[Zhu et~al\mbox{.}(2022b)]%
        {zhu2022memory}
\bibfield{author}{\bibinfo{person}{Yongchun Zhu}, \bibinfo{person}{Qiang Sheng}, \bibinfo{person}{Juan Cao}, \bibinfo{person}{Qiong Nan}, \bibinfo{person}{Kai Shu}, \bibinfo{person}{Minghui Wu}, \bibinfo{person}{Jindong Wang}, {and} \bibinfo{person}{Fuzhen Zhuang}.} \bibinfo{year}{2022}\natexlab{b}.
\newblock \showarticletitle{Memory-Guided Multi-View Multi-Domain Fake News Detection}.
\newblock \bibinfo{journal}{\emph{IEEE Transactions on Knowledge and Data Engineering}} (\bibinfo{year}{2022}).
\newblock


\end{thebibliography}

\appendix

\section{Discussion: Effect of Reframings on LLM Style Robustness}
\label{sec:reframing-llm-robustness}

This section investigates the effects of incorporating our style-diverse news reframings (Section \ref{sec:llm-reframe}) on LLM style robustness. Following the experimental setup for evaluating style robustness (Section \ref{sec:sheepdog-experiments}), \textbf{\textit{we report results on adversarial set A (formulation described in Table \ref{tab:adv-setup}) unless otherwise specified}}. We explore two methods for adapting LLMs to diverse news styles: \textbf{\textit{(1)}} \textbf{in-context learning} and \textbf{\textit{(2)}} \textbf{fine-tuning}, with prompt templates and LLM configurations detailed in Appendix \ref{sec:llm-detection-prompt}. To mitigate randomness, we present averaged metrics from three runs of each approach.

\textbf{\underline{In-Context Learning.}} Table \ref{tab:reframe-icl} compares a zero-shot GPT-3.5 detector with the following three in-context learning variants: \textbf{\textit{(1)}} \textbf{GPT-3.5+ICL-2}, which incorporates 2 randomly selected in-context samples (1 real and 1 fake). \textbf{\textit{(2)}} \textbf{GPT-3.5+ICL-2-R}, which incorporates the same 2 in-context samples as GPT-3.5+ICL-2, enriched with style-diverse reframings. For each in-context sample, we randomly select one reliable-style reframing and one unreliable-style reframing. \textbf{\textit{(3)}} \textbf{GPT-3.5+ICL-4}, which incorporates 4 randomly selected in-context samples (2 real and 2 fake). 
\begin{table}[ht!]
\caption{GPT-3.5 in-context learning performance.}
{
 \begin{tabular}{lccc} \toprule
 \textbf{Method} &  \textbf{PolitiFact} & \textbf{GossipCop} & \textbf{LUN} \\ 
 \toprule
 GPT-3.5  & 42.13  & 39.59 & 59.63\\
 $\;$ + ICL-2  & 38.57 & 38.88 & 57.51 \\
 $\;$ + ICL-2-R  & 41.76 & \textbf{41.83} & \textbf{60.29}\\
 $\;$ + ICL-4  & \textbf{42.35} & 41.47 & 60.25\\

 \bottomrule
\end{tabular} 
}
 \label{tab:reframe-icl}
\end{table}

Results in Table \ref{tab:reframe-icl} indicate three key findings: \textbf{\textit{(1)}} \textit{LLMs benefit from more in-context samples}, as GPT-3.5+ICL-4 outperforms GPT-3.5+ICL-2. \textbf{\textit{(2)}} \textit{Incorporation of news reframings effectively enhances LLM style robustness}, as shown by the consistent gains of GPT-3.5+ICL-2-R over GPT-3.5+ICL-2. \textbf{\textit{(3)}} \textit{On the fake news detection task, LLM in-context learning does not offer clear benefits over LLM zero-shot.} The long sequence lengths of news articles limit the number of in-context demonstrations, resulting in insufficient context and overly lengthy prompts that LLMs struggle to process.

\textbf{\underline{Fine-Tuning.}} Table \ref{tab:reframe-ft} compares a zero-shot LLaMA2-13B detector with the following two fine-tuned variants: \textbf{\textit{(1)}} \textbf{LLaMA2-13B+FT}, which fine-tunes LLaMA2 using the original training articles. \textbf{\textit{(2)}} \textbf{LLaMA2-13B+FT-R}, which fine-tunes LLaMA2 using the original training articles and their style-diverse reframings. For each article, we randomly incorporate one reliable-style reframing and one unreliable-style reframing as fine-tuning data.

\begin{table}[ht!]
\caption{LLaMA2-13B fine-tuning performance.}
\small
{
 \begin{tabular}{lccc} \toprule
 \textbf{Method} &  \textbf{PolitiFact} & \textbf{GossipCop} & \textbf{LUN} \\ 
 \toprule
 LLaMA2-13B  & 33.24 & 25.79 & 32.64\\
 $\;$ + FT  & 36.96 & 47.83  & 34.31 \\
 $\;$ + FT-R  & \textbf{44.44} & \textbf{56.06} & \textbf{42.22} \\
 \bottomrule
\end{tabular} 
}
 \label{tab:reframe-ft}
\end{table}

Both LLaMA2 variants are fine-tuned using QLoRA \cite{dettmers2023qlora} for 1 epoch with a batch size of $16$ and a learning rate of $1\times10^{-4}$. From Table \ref{tab:reframe-ft}, we observe: \textbf{\textit{(1)}} \textit{Fine-tuning generally enhances task-specific LLM capabilities on fake news detection}, as indicated by consistent improvements of LLaMA2-13B+FT over LLaMA2-13B. \textbf{\textit{(2)}} \textit{Incorporating style-diverse news reframings as fine-tuning samples further improves style robustness}, as indicated by consistent improvements of LLaMA2-13B+FT-R over LLaMA2-13B+FT. \textbf{\textit{(3)}} \textit{LLaMA2-13B+FT-R still falls short compared to GPT-3.5 on the LUN adversarial test set and performs worse than task-specific detectors and fully fine-tuned LMs in Table \ref{tab:adv-performance}}. This suggests the need for larger, task-specific corpora for fully adapting LLMs to fake news detection, further highlighting our motivation for leveraging LLM general-purpose capabilities in a zero-shot manner, which are reasonably strong and more readily available.

\section{LLM Prompting Configurations}
\label{sec:llm-prompt-detail}

\subsection{LLM Baselines}
\label{sec:llm-detection-prompt}

In section \ref{sec:baselines}, we select three representative LLMs as fake news detection baselines: \textbf{GPT-3.5} \cite{openai2022chatgpt} (model name: \texttt{gpt-3.5-turbo-0301}), \textbf{InstructGPT} \cite{ouyang2022training} (model name: \texttt{gpt-3.5-turbo-instruct}), and \textbf{LLaMA2-13B} \cite{touvron2023llama} (model name: \texttt{Llama-2-13b-chat-hf}). 

For GPT-3.5 and InstructGPT, we use their APIs from OpenAI, and set the temperature to $0$ for stable veracity predictions. For LLaMA2-13B, we employ model weights from HuggingFace Transformers \cite{wolf20transformers} version 4.31.0, with \texttt{do\_sample} set to False for greedy decoding. All models adopt the following prompt for zero-shot fake news detection:

\begin{tcolorbox}[colback=black!2!white,colframe=white!50!black,boxrule=0.5mm]
\small
  \textbf{Question}: Does the following contain real or fake news (or information)? Answer in one word with either `Real' or `Fake', then explain why. [news article]\\
  \textbf{Answer}: [starts with a predicted veracity label]
  
\end{tcolorbox}

\subsection{Obtaining SheepDog's News Reframings}
\label{sec:llm-reframe-prompt}

Recall from Section \ref{sec:llm-reframe} that we generate reliable-style reframings and unreliable-style reframings with GPT-3.5 for each labeled news article. Among the four prompt templates presented below, we use the first two templates for reliable-style reframings, and the other two for unreliable-style reframings. 

During reframing generation, we set the temperature to $0.7$ and limit the maximum number of response tokens to $512$. See Table \ref{tab:news-reframing-example} for a detailed reframing example.

\begin{tcolorbox}[colback=black!2!white,colframe=white!50!black,boxrule=0.5mm]
\small

  Rewrite the following article in an \textit\textbf{{objective and professional}} tone: [news article]
\end{tcolorbox}

\begin{tcolorbox}[colback=black!2!white,colframe=white!50!black,boxrule=0.5mm]
\small

  Rewrite the following article in a \textit\textbf{{neutral}} tone: [news article]
\end{tcolorbox}

\begin{tcolorbox}
[colback=black!2!white,colframe=white!50!black,boxrule=0.5mm]
\small

  Rewrite the following article in an \textit\textbf{{emotionally triggering}} tone: [news article]
\end{tcolorbox}

\begin{tcolorbox}[colback=black!2!white,colframe=white!50!black,boxrule=0.5mm]
\small

  Rewrite the following article in a \textit\textbf{{sensational}} tone: [news article]
\end{tcolorbox}

\subsection{Obtaining SheepDog's Veracity Attributions}
\label{sec:llm-attribution-prompt}

In Section \ref{sec:llm-attributions}, we use an LLM to elicit auxiliary content-centric information directly related to news veracity. Specifically, we prompt the LLM to provide explanatory outputs for each fake news article in the training set, based on predefined content-focused rationales aimed at debunking fake news.

We draw from existing literature, which commonly emphasizes two key aspects: \textbf{\textit{(1)}} \textbf{sources of information} \cite{jung2020caution,zhang2018structured}, and \textbf{\textit{(2)}} \textbf{correctness of news content}  \cite{allcott2017social,shu2017fake}. Inspired by these insights, we devise four rationales:

\begin{itemize}[leftmargin=*]
    \item \textbf{Source}-related rationales: \textbf{\textit{[A]}} lack of credible sources; and \textbf{\textit{[B]}} inconsistencies with reputable sources. These rationales are informed by prior research emphasizing the importance of reputable sources in validating conclusions \cite{zhang2018structured} and correcting false information \cite{jung2020caution}.
    
    \item \textbf{Content}-related rationales:  \textbf{\textit{[C]}} false or misleading information; and \textbf{\textit{[D]}} biased opinion.  Fake news, defined as intentionally and verifiably false information \cite{allcott2017social,shu2017fake}, often manipulates biased opinions to exploit cognitive vulnerabilities in news consumers. Consequently, we devise \textbf{\textit{[C]}} in alignment with the definition of fake news, and \textbf{\textit{[D]}} to represent a common strategy employed by fake news producers.

\end{itemize}

Our four rationales cover the key aspects of source- and content-related indicators in debunking fake news. We utilize the following prompt to obtain content-focused veracity attributions, and set the temperature to $0$ to ensure stability. An example is presented in Table \ref{tab:veracity-attribution-example}. 
\begin{tcolorbox}[colback=black!2!white,colframe=white!50!black,boxrule=0.5mm]
\small

  \textbf{Article}: [fake news article]\\
  \textbf{Question}: Which of the following problems does this article have? Lack of credible sources, False or misleading information, Biased opinion, Inconsistencies with reputable sources. If multiple options apply, provide a comma-separated list ordered from most to least related. Answer ``No problems'' if none of the options apply.
\end{tcolorbox}

\section{Analysis on Content Consistency of LLM-Empowered News Reframings}
\label{sec:content-consistency}

LLM-empowered news reframings play a key role in achieving style robustness. Table \ref{tab:news-reframing-example} provides a comparison between a news article and its reliable-style reframing, illustrating concretely the impressive capability of LLMs to introduce different tones while preserving the news content.

To offer a more comprehensive assessment of content consistency between the original news article and its reframings, we investigate \textit{claim entailment} for a straightforward and quantifiable estimation. Specifically, the original article should ideally entail the central factual claims within its reframing, and vice versa. 

Given that news articles exhibit significantly longer sequences and more complex logical structures compared to the sentence pairs in natural language inference (NLI) benchmarks (e.g., SNLI \cite{bowman2015large} and MultiNLI \cite{williams2018mnli}), we opt against utilizing NLI models pre-trained on these benchmarks. Instead, \textit{we instruct a GPT-3.5 model to extract claims and infer claim entailment}. The prompt template for claim extraction is as follows:

\begin{tcolorbox}[colback=black!2!white,colframe=white!50!black,boxrule=0.5mm]
\small

 Extract and summarize the central factual claim in the following article. Article: [news article A]. Claim: 
\end{tcolorbox}
Querying the LLM yields a succinct summarization of the article's central factual claim, typically composed of several sub-claims. An example response is provided as follows:
\begin{tcolorbox}[colback=black!2!white,colframe=white!50!black,boxrule=0.5mm]
\small

 \textbf{Response Example for Claim Extraction:} Financial experts are concerned about the negative impact of China's undervalued yuan on both Asia and the United States. They are calling on regional governments and the Group of 20 leaders to take action to prevent potential currency and trade wars. The experts emphasize the need for neighboring countries to urge China to relax its exchange rate controls in order to address the global current account imbalance. They also highlight the adverse effects of US monetary easing and China's low exchange rate on emerging market economies. The experts are urging the G20 to address the currency problem at its upcoming summit in South Korea and to oppose unilateral devaluation moves and support currency stability. 
\end{tcolorbox}

With the extracted claim, we predict claim entailment using the following prompt:
\begin{tcolorbox}[colback=black!2!white,colframe=white!50!black,boxrule=0.5mm]
\small

 \textbf{Question}: Does the following article entail the claim: [claim extracted from news article A]? Answer in one word with either `Yes' or `No'. Article: [news article B]. 
\end{tcolorbox}

While precise entailment assessment requires detailed analysis of each article's nuances and specific claim wordings, evaluating the proportion of article pairs with predicted claim entailment provides a general estimation of content consistency. To ensure comprehensive coverage, we analyze two larger-scale benchmark datasets: GossipCop and LUN (dataset statistics in Table \ref{tab:ds-stats}).
\begin{table}[ht!]
    \centering
    \caption{Claim entailment (\%) between original news articles and their objective-style reframings.}
    \label{tab:claim-entailment}
\resizebox{\columnwidth}{!}{
    \begin{tabular}{lcc}
        \toprule
        \textbf{Dataset} & original-entail-objective & objective-entail-original \\
        \midrule
        GossipCop& 86.20 & 89.17 \\
        LUN& 89.22 & 87.53 \\
        \bottomrule
    \end{tabular}}
    
\end{table}

As shown in Table \ref{tab:claim-entailment}, the proportions of claim entailment range from 86.20\% to 89.22\% across both datasets, suggesting a reasonably high consistency between the central factual claims presented in the original news articles and their reframings. This consistency further validates the effectiveness of our reframing approach in preserving the core content of news while injecting stylistic variations.

\begin{table*}[ht]
    \centering
    \caption{An example of GPT3.5-generated reliable-style news reframing, truncated into sentences for a clearer comparison.}
    \label{tab:news-reframing-example}
    \begin{tabular}{p{0.45\linewidth} p{0.45\linewidth}}
        \toprule
        \textbf{PROMPT}: Rewrite the following article in an objective and professional tone: & \textbf{RESPONSE}: \\
        \midrule
        \vspace{1pt}BREAKING: Obamas Daughter BUSTED By The Cops For Doing This & \vspace{1pt}Malia Obama's Recent Activities Raise Concerns \\
        \vspace{1pt} At the Democratic National Convention, Michelle Obama tried to use her daughters as shining examples of just how far America has come. &
        \vspace{1pt} Recently, reports have emerged regarding the behavior of Malia Obama, daughter of former President Barack Obama and Michelle Obama. At the Democratic National Convention, Michelle Obama showcased her daughters as symbols of progress in America. \\
        \vspace{1pt} 
        But the fact is that the young Obamas are pushing the limits as any teenager does and probably even further. & \vspace{1pt}However, it appears that the young Obamas, like many teenagers, are testing boundaries and possibly going beyond them. \\
        \vspace{1pt}According to new reports, Malia Obama was spotted at a party in Marthas Vineyard recently. It was eventually broken up by police because of noise complaints.  & \vspace{1pt} 
        According to sources, Malia Obama was seen at a gathering in Marthas Vineyard that was eventually disbanded by law enforcement due to noise complaints. \\
        \vspace{1pt} 
        The first daughter was immediately rushed from the party by Secret Service before the cops arrived. & \vspace{1pt} The first daughter was promptly escorted away from the scene by the Secret Service before the arrival of the police.\\
        \vspace{1pt} The West Tisbury police declined to confirm if Malia was there, citing national security. & \vspace{1pt}West Tisbury police have neither confirmed nor denied Malia's presence at the event, citing national security concerns. \\
        \vspace{1pt}But the truth surfaced anyway. & \vspace{1pt} Nevertheless, the truth regarding her attendance eventually surfaced. \\
        \vspace{1pt} This is just one incident in a string of wild behavior on Malias part. & \vspace{1pt} This incident is just one in a series of instances where Malia's behavior has attracted attention.\\
        \vspace{1pt} Just a month ago the first daughter came under fire when photos surfaced of her smoking marijuana at a concert. &
        \vspace{1pt} Not long ago, she faced criticism when photos emerged showing her allegedly smoking marijuana at a concert. \\
        \vspace{1pt} Malia is clearly using her gap year before Harvard to party and experiment. & \vspace{1pt} It appears that Malia is utilizing her gap year before attending Harvard University as a period for socializing and experimentation. \\
        \vspace{1pt} Later, Malia and Barack were spotted strolling around a pool together having a chat. & \vspace{1pt} Later on, both Malia and Barack Obama were observed engaging in a conversation while walking near a pool. \\ 
        \vspace{1pt} Were sure he had a few choice words for her. & \vspace{1pt} It can be assumed that the former president had some words of advice or guidance for his daughter. \\
        \vspace{1pt} Clearly, a lack of respect for the law runs deep in this family. &
        \vspace{1pt} It is evident that a lack of regard for the law seems to be a recurring issue within this family.  In conclusion, recent events raise concerns about the behavior of Malia Obama. These incidents, including her presence at a party disbanded by the police and alleged marijuana use, suggest a period of exploration and a certain disregard for legal boundaries. \\

        \bottomrule
    \end{tabular}
\end{table*}

\begin{table*}[ht]

    \centering
    \caption{An example of content-focused veracity attributions extracted from GPT3.5, given a fake news article as input.}
    \label{tab:veracity-attribution-example}
    \resizebox{1.0\textwidth}{!}{
    \begin{tabular}{p{18cm}}
        \toprule
        \\ \underline{\textbf{PROMPT}}: \textbf{Article:} BREAKING: Obamas Daughter BUSTED By The Cops For Doing This At the Democratic National Convention, Michelle Obama tried to use her daughters as shining examples of just how far America has come. But the fact is that the young Obamas are pushing the limits as any teenager doesand probably even further. According to new reports, Malia Obama was spotted at a party in Marthas Vineyard recently. It was eventually broken up by police because of noise complaints. The first daughter was immediately rushed from the party by Secret Service before the cops arrived. The West Tisbury police declined to confirm if Malia was there, citing national security. But the truth surfaced anyway. This is just one incident in a string of wild behavior on Malias part. Just a month ago the first daughter came under fire when photos surfaced of her smoking marijuana at a concert. Malia is clearly using her gap year before Harvard to party and experiment. Later, Malia and Barack were spotted strolling around a pool together having a chat. Were sure he had a few choice words for her. Clearly, a lack of respect for the law runs deep in this family.  \\
        \vspace{1pt}\textbf{Question: Which of the following problems does this article have? Lack of credible sources, False or misleading information, Biased opinion, Inconsistencies with reputable sources. If multiple options apply, provide a comma-separated list ordered from most to least related. Answer "No problems" if none of the options apply.} 
\\ \\
\midrule
        \\ \underline{\textbf{RESPONSE}}:  False or misleading information, Biased opinion.

\\

        \bottomrule
    \end{tabular}
    }
\end{table*}

\end{document}